\pdfoutput=1

\documentclass[11pt]{article}

\usepackage[final]{acl}

\usepackage{times}
\usepackage{latexsym}

\usepackage[T1]{fontenc}

\usepackage[utf8]{inputenc}

\usepackage{microtype}

\usepackage{inconsolata}
\usepackage{CJKutf8}
\usepackage{tcolorbox}
\usepackage{amsmath}
\usepackage{amsfonts}
\usepackage{makecell}
\usepackage{multirow}
\usepackage{booktabs}
\usepackage{graphicx}
\usepackage{subfigure}
\usepackage{multicol}
\usepackage{microtype}
\usepackage{listings}
\usepackage{subcaption}
\usepackage{pifont}
\usepackage{bbding}
\usepackage{xcolor}
\usepackage{colortbl}
\usepackage{amssymb}
\usepackage{amsfonts}
\usepackage{float}
\usepackage{mdframed}
\usepackage{soul}
\usepackage{ulem}
\usepackage[inline]{enumitem}
\usepackage{enumitem}
\usepackage{algorithm}
\usepackage{algorithmicx}
\usepackage{algcompatible}
\usepackage{nccmath}   
\usepackage{wrapfig} 
\usepackage{lipsum} 
\usepackage{tabularray}
\tcbuselibrary{breakable}
\definecolor{avgcol}{RGB}{255,252,223}
\definecolor{succcol}{RGB}{230,230,255}
\definecolor{oursgray}{RGB}{245,245,245}
\newcolumntype{Y}{>{\columncolor{avgcol}}c}
\newcolumntype{P}{>{\columncolor{succcol}}c}
\title{SGR-OPSD: Learning from Subgoal Plans with Graph-Regularized Credit Assignment for Agentic Reinforcement Learning}

\author{
 \textbf{Shu Liu\textsuperscript{1,2\thanks{\ \ This work was done during an internship at Meituan.}}},
 \textbf{Yang Chong\textsuperscript{2}},
 \textbf{WenShi Chen\textsuperscript{2}},
 \textbf{Man Lan\textsuperscript{1}\thanks{\ \ Corresponding author.}}
\\
 \textsuperscript{1}School of Computer Science and Technology, East China Normal University
 \\
 \textsuperscript{2}MeiTuan
\\
    \texttt{shuliu@stu.ecnu.edu.cn, mlan@cs.ecnu.edu.cn} \\
}

\begin{document}
\maketitle

\begin{abstract}
OPSD provides dense supervision for agentic reinforcement learning through privileged information feedback. However, existing privileged information often lacks explicit task-progression planning, while teacher--student log-probability gaps reflect shifts in model preference rather than true task progress. To address these limitations, we propose Subgoal-Guided Graph-Regularized OPSD (\textbf{SGR-OPSD}), a framework that combines subgoal supervision with interaction-derived evidence for turn-level credit assignment. Specifically, \textbf{SGR-OPSD} uses LLM-generated subgoal plans to provide dense action-level guidance, converting them into advantage weights via a constrained, KL-regularized objective. To ground credit in observed outcomes, we construct a Subgoal Progress Graph that estimates empirical returns from sampled transition frequencies. The graph compares empirical action returns against the state average to derive preferences that calibrate the teacher's credit weights. Evaluated on ALFWorld, WebShop, and Search-QA across two model scales (Qwen2.5-3B and 7B), \textbf{SGR-OPSD} consistently outperforms GRPO and strong self-distillation baselines—improving success rates over GRPO by up to 7.6 percentage points and reaching \textbf{85.8\%} on ALFWorld with Qwen2.5-7B. Ablation studies confirm that subgoal guidance and graph-based weight calibration are key to these performance gains.
\end{abstract}

\section{Introduction}
\label{sec:intro}
\begin{figure}[ht!]
    \centering
    \includegraphics[width=0.5\textwidth]{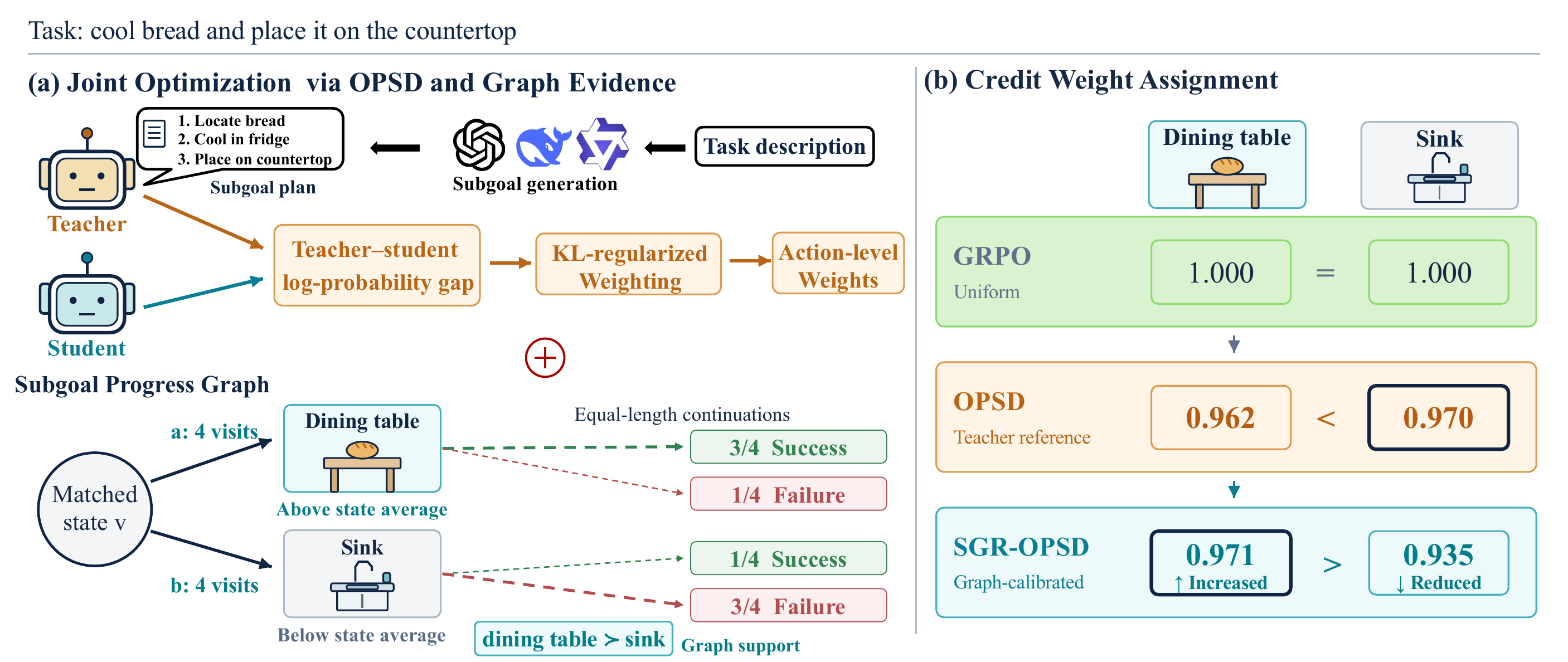}
    \caption{Overview of \textbf{SGR-OPSD}. (a) Subgoal-guided OPSD produces credit weights, which are jointly calibrated using subgoal progress graph evidence. (b) The figure shows a real example in which calibration shifts favorable credit from the sink to the dining table. 
    }
    \label{fig:credit-calibration}
\end{figure}

Reinforcement learning (RL) enables LLM-based agents to solve complex tasks through multi-turn interactions, ranging from household manipulation to online shopping~\citep{shridhar2020alfworld,yao2022webshop}. Effective training requires fine-grained supervision to evaluate intermediate decisions within a trajectory. On-Policy Self-Distillation (OPSD)~\citep{zhao2026selfdistilledreasoner} provides dense feedback by conditioning a teacher model on privileged training-time information to score student-sampled actions~\citep{lv2026pcsd,lan2026trust}. This feedback complements the sparse outcome rewards used in methods like Group Relative Policy Optimization (GRPO)~\citep{shao2024deepseekmath}, enabling the student to internalize inference-unavailable guidance. OPSD thus offers a promising framework for dense supervision in agentic RL.

Existing methods supply privileged information through trajectory-derived skills or reference answers, integrating teacher feedback via advantage reweighting or auxiliary distillation objectives~\citep{wang2026skillsd,yang2026rlsd,lu2026sdar}. AgentOPSD further accumulates likelihood evidence across turns, while TASPO maps action-level support into bounded advantage weights~\citep{wang2026agentopsd,yang2026taspo}. However, two key gaps remain between such supervision and true task progress. First, privileged guidance often lacks an explicit, ordered plan for the task at hand. For \textit{``cool bread and place it on the countertop''}, a structured plan—locate the bread, cool it, then place it—makes the required progression explicit. Second, teacher--student log-probability gaps measure shifts in model preference rather than progress made through real interactions; even plan-conditioned feedback can incorrectly favor actions that stray from successful continuations. Thus, effective credit assignment requires both task-specific planning information and a mechanism that grounds its influence in execution evidence. To bridge these gaps, our approach uses subgoal plans to structure teacher guidance and uses evidence from the Subgoal Progress Graph to calibrate the resulting credit.

We propose \textbf{SGR-OPSD}, illustrated in Figure~\ref{fig:credit-calibration}. An LLM generates a fixed subgoal plan from the task description alone, providing the scoring teacher with explicit task-progression guidance. We derive a constrained KL-regularized rule that converts action-level teacher--student likelihood evidence into bounded exponential weights on GRPO advantages. The rule accounts for advantage signs and enforces a fixed token-weighted budget within each trajectory, establishing a teacher reference for turn-level credit. To calibrate these weights, we construct a \emph{Subgoal Progress Graph} from student rollouts and derive action preferences from empirical $Q-V$: each action's discounted return relative to the visit-weighted state average. A joint objective incorporates these preferences while regularizing the allocation toward the teacher reference, connecting plan-guided supervision with evidence of task progress. The trained student requires neither the plan nor the graph at inference. We evaluate \textbf{SGR-OPSD} on ALFWorld, WebShop, and Search-QA with Qwen2.5-3B and Qwen2.5-7B. On the 7B model, it reaches \textbf{85.8\%} success on ALFWorld and \textbf{78.2\%} on WebShop, improving over GRPO by 7.6 percentage points on both benchmarks.

Our contributions are as follows:

\begin{itemize}[leftmargin=*]
    \item We propose leveraging subgoal plans as privileged supervision and derive a constrained KL-regularized formulation to allocate turn-level credit via bounded exponential weights.
    \item We develop a Subgoal Progress Graph that derives action preferences from empirical return estimates to calibrate teacher-reference weights, while preserving advantage signs and a fixed token-weighted budget per trajectory.
    \item Extensive experiments across three environments and two model scales demonstrate the superiority of \textbf{SGR-OPSD}, outperforming GRPO by up to 7.6 percentage points and achieving an \textbf{85.8\%} success rate on ALFWorld.
\end{itemize}

\section{Related Works}
\label{sec:relatedworks}
\subsection{On-Policy Self-Distillation for LLMs}
Self-distillation derives the teacher from the student and conditions it on privileged training-time information~\cite{wang2026skillsd,wang2026agentopsd}. Existing works integrate diverse privileged signals via distinct mechanisms: Skill-SD leverages trajectory-derived skills~\cite{wang2026skillsd}, RLSD scales token updates using log-probability gaps between an answer-conditioned teacher and the student~\cite{yang2026rlsd}, and SDAR gates SkillBank guidance alongside RL~\cite{lu2026sdar}. However, these privileged signals often contain irrelevant noise and lack explicit task-progress planning. To address this, \textbf{SGR-OPSD} uses a fixed, LLM-generated subgoal plan derived solely from the task description as privileged information and applies a constrained KL-regularized objective to transform action-level evidence into bounded turn weights.

\subsection{Credit Assignment for Multi-Turn Agents}
While GRPO provides trajectory-level advantages under outcome rewards, it fails to distinguish the contributions of individual turns~\citep{shao2024deepseekmath}. Recent lines of work refine this coarse outcome signal: AgentOPSD accumulates privileged likelihood evidence via recursive belief updates~\citep{wang2026agentopsd}, TASPO uses action-level support for bounded advantage reweighting~\citep{yang2026taspo}, and AlignOPSD aligns sibling contexts to allocate credit across spans and turns~\citep{chen2026alignopsd}. Although such teacher likelihood preferences offer dense feedback, they lack explicit measurements of task progress. To bridge this gap, \textbf{SGR-OPSD} constructs a Subgoal Progress Graph from student rollouts, leveraging graph-derived action comparisons to calibrate turn-level credit from a subgoal-conditioned teacher.

\section{Method}
\label{sec:method}
\begin{figure*}[ht!]
    \centering
    \includegraphics[width=1\textwidth]{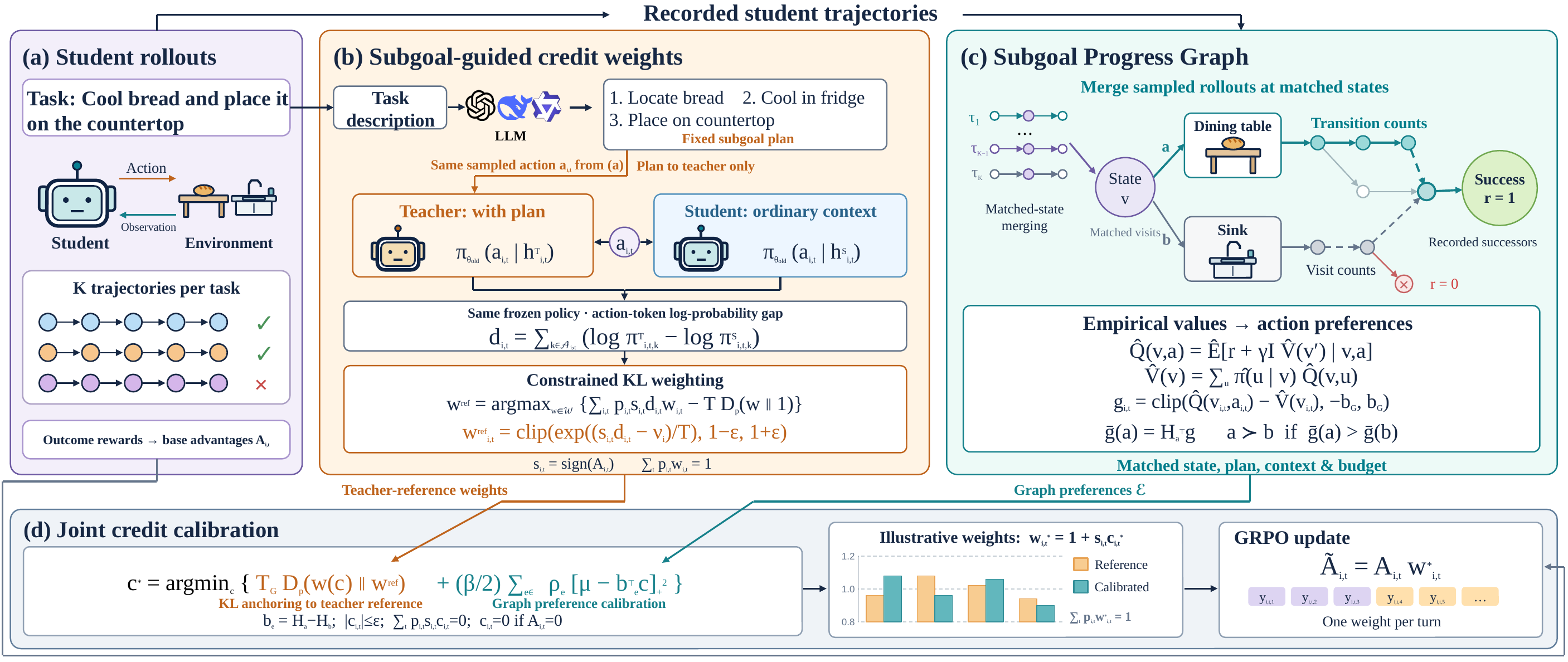}
    \caption{Overview of \textbf{SGR-OPSD}. Subgoal plans guide teacher feedback, which constrained KL weighting converts into turn-level credit weights. The Subgoal Progress Graph uses student trajectories to derive action preferences and calibrate these weights. The calibrated weights $w_i^*$ redistribute GRPO credit while preserving advantage signs and a fixed token-weighted budget.
    }
    \label{fig:method-overview}
\end{figure*}

\textbf{SGR-OPSD} combines plan-conditioned teacher preferences with rollout-derived outcome evidence for turn-level credit assignment. Constrained KL weighting converts teacher feedback into reference weights, which the Subgoal Progress Graph calibrates through action comparisons (Figure~\ref{fig:method-overview}).

\subsection{Background}
\label{sec:background}
We consider multi-turn tasks where an agent interacts with an environment to complete an instruction $x$. Given a task $x$, the rollout policy $\pi_{\theta_{\mathrm{old}}}$ samples $K$ trajectories $\{\tau_i\}_{i=1}^K$. Each trajectory $\tau_i$ consists of $N_i$ turns indexed by $t$, with $k$ indexing individual tokens within a response. At turn $t$, given context $h_{i,t}^S$, the agent generates a response $y_{i,t}$ containing action $a_{i,t}$ and receives the next observation.

Outcome-based policy optimization assigns each turn a base advantage $A_{i,t}$, where positive values reinforce the sampled turn and negative values discourage it. Self-distillation supplements this sparse outcome feedback by re-scoring student-sampled trajectories using privileged information. Both scoring branches share the pre-update parameters $\theta_{\mathrm{old}}$, so their likelihood difference reflects the effect of privileged information.

\subsection{Subgoal-Guided Credit Weights}
\label{sec:subgoal}
\label{sec:teacher}
\textbf{Subgoal-conditioned supervision.}
Subgoal plans make the intermediate requirements of task completion explicit. We employ a fixed, ordered plan $z(x)$, generated by DeepSeek-V3.1~\citep{deepseekai2024deepseekv3technicalreport} solely from the task instruction $x$, as privileged information. $z(x)$ combines task-specific intermediate goals and ordering constraints without requiring prior task interactions or successful trajectories (Appendix~\ref{app:subgoal-generation}). Figure~\ref{fig:method-overview}(b) illustrates how this plan conditions teacher scoring to yield subgoal-guided teacher--student log-likelihood gaps.  

Specifically, the teacher branch receives the privileged context $h_{i,t}^T$ augmented with $z(x)$, whereas the student operates on the standard context $h_{i,t}^S$. Token probabilities for branch $B \in \{T, S\}$ are defined as $\pi^B_{i,t,k} = \pi_{\mathrm{old}}(y_{i,t,k} \mid h_{i,t}^B, y_{i,t,<k})$, with both branches retaining identical response prefixes $y_{i,t,<k}$. Summing the token-level log-probability differences over the token positions $\mathcal{A}_{i,t}$ of the executed action yields the action-level gap:
\begin{equation}
    d_{i,t} = \sum_{k \in \mathcal{A}_{i,t}} \bigl(\log \pi^T_{i,t,k} - \log \pi^S_{i,t,k}\bigr).
    \label{eq:gap}
\end{equation}
The gap $d_{i,t}$ measures how the plan changes the action's log-likelihood, providing dense action-level evidence for credit assignment.

\textbf{Constrained KL weighting.}
Subgoal support should strengthen positive updates and soften negative ones. We derive turn weights by balancing signed teacher evidence against KL regularization toward unit weights:
\begin{equation}
 w^{\mathrm{ref}}=\arg\max_{w\in\mathcal W}
 \bigl\{\mathcal U(w)-T D_p(w\|\mathbf 1)\bigr\},
\label{eq:teacher}
\end{equation}
where $w=(w_{i,t})_{i,t}$ collects the turn weights and $w^{\mathrm{ref}}$ is their optimized teacher reference. The feasible set $\mathcal W$ bounds each weight by $1-\epsilon\leq w_{i,t}\leq1+\epsilon$ ($0\leq\epsilon<1$) to limit redistribution, fixes $w_{i,t}=1$ when $A_{i,t}=0$, and imposes a token-weighted budget for each trajectory:
\begin{equation}
 \sum_{t=1}^{N_i}p_{i,t}w_{i,t}=1,
 \qquad p_{i,t}=\frac{m_{i,t}}{\sum_{r=1}^{N_i}m_{i,r}},
\label{eq:mass}
\end{equation}
where $m_{i,t}$ counts effective response tokens in turn $t$. Thus, $p_{i,t}$ weights turns by their token share, and the budget redistributes emphasis while keeping the weighted mean at one.
Within this budget, the signed teacher-support term is
\begin{equation}
 \mathcal U(w)=\sum_{i,t}p_{i,t}s_{i,t}d_{i,t}w_{i,t},
\end{equation}
where $d_{i,t}$ is the action-level gap in Equation~\ref{eq:gap} and $s_{i,t}=\operatorname{sign}(A_{i,t})$. The sign factor makes stronger teacher support favor larger weights for positive advantages and smaller weights for negative ones.

The KL term limits departures from unit weights, with $T>0$ controlling its strength. We use the token-weighted generalized KL divergence
\begin{equation}
 \resizebox{0.86\columnwidth}{!}{$\displaystyle
 D_p(u\|v)=\sum_{i,t}p_{i,t}\left(u_{i,t}\log\frac{u_{i,t}}{v_{i,t}}-u_{i,t}+v_{i,t}\right),
 $}
\label{eq:weighted-kl}
\end{equation}
where $u$ is a positive weight vector and $v$ its positive reference, with entries $u_{i,t}$ and $v_{i,t}$. In Equation~\ref{eq:teacher}, these are $w$ and the all-ones vector $\mathbf 1$, respectively. For $A_{i,t}\neq0$, this objective yields
\begin{equation}
 \resizebox{0.86\columnwidth}{!}{$\displaystyle
 w_{i,t}^{\mathrm{ref}}=\operatorname{clip}\!\left(
 \exp\!\left(\frac{s_{i,t} d_{i,t}-\nu_i}{T}\right),
 1-\epsilon,1+\epsilon\right),
 $}
\label{eq:tilt}
\end{equation}
where $\nu_i$ is the trajectory-specific multiplier enforcing the budget (Appendix~\ref{app:implementation}). These bounded weights provide the reference for calibration by the Subgoal Progress Graph.

\begin{table*}[t]
\centering
\small
\definecolor{SGRrow}{RGB}{242,244,247}
\definecolor{SGRavg}{RGB}{237,243,251}
\definecolor{SGRsearch}{RGB}{218,240,229}
\definecolor{SGRsucc}{RGB}{237,247,239}
\setlength{\tabcolsep}{3.4pt}
\renewcommand{\arraystretch}{1.08}
\resizebox{\textwidth}{!}{%
\begin{tabular}{@{}l *{6}{c} >{\columncolor{SGRavg}}c *{7}{c} >{\columncolor{SGRsearch}}c c >{\columncolor{SGRsucc}}c@{}}
\toprule
& \multicolumn{7}{c}{\textbf{ALFWorld}} & \multicolumn{8}{c}{\textbf{SearchQA}} & \multicolumn{2}{c}{\textbf{WebShop}} \\
\cmidrule(lr){2-8}\cmidrule(lr){9-16}\cmidrule(lr){17-18}
\textbf{Method} & Pick & Look & Clean & Heat & Cool & Pick2 & Avg
& NQ & Triv & Pop & Hotp & 2Wk & MuS & Bam & Avg & Score & Succ. \\
\midrule
\multicolumn{18}{l}{\textit{Closed-Source Models}} \\
\midrule
Qwen3.5-Plus
& 82.9 & 30.8 & 66.7 & 6.3 & 44.0 & 66.7 & 49.5
& 28.5 & 55.1 & 35.0 & 28.6 & 37.2 & 23.5 & 42.8 & 35.8
& 9.8 & 9.5 \\
DeepSeek-V3.1
& 91.4 & 69.2 & 48.1 & 50.0 & 32.0 & 54.2 & 57.5
& 20.6 & 64.0 & 35.6 & 35.4 & 31.0 & 18.6 & 54.4 & 37.1
& 10.2 & 8.4 \\
\midrule
\multicolumn{18}{l}{\textit{Qwen2.5-3B-Instruct}} \\
\midrule
Vanilla
& 44.4 & 11.1 & 6.2 & 15.4 & 28.6 & 12.5 & 19.7
& 24.6 & 48.1 & 31.0 & 26.3 & 25.3 & 7.2 & 21.5 & 26.3
& 6.7 & 0.8 \\
GRPO
& 91.4 & 66.6 & 75.3 & 58.3 & 61.3 & 62.5 & 69.3
& \underline{45.3} & \underline{61.2} & 44.9 & \underline{40.1} & 38.1 & 15.1 & 31.2 & 39.4
& 78.5 & 63.1 \\
Skill-Prompt
& 51.7 & \underline{66.7} & 48.4 & 0.0 & 4.3 & 10.0 & 30.2
& 23.7 & 46.2 & 30.6 & 24.4 & 22.1 & 7.5 & 12.5 & 23.9
& 4.9 & 1.2 \\
OPSD
& 48.8 & 41.7 & 16.7 & 0.0 & 15.8 & 16.7 & 23.3
& 0.1 & 0.1 & 0.1 & 0.0 & 0.0 & 0.0 & 0.0 & 0.0
& 11.3 & 3.1 \\
GRPO+OPSD
& 95.2 & 59.2 & \underline{84.0} & 60.4 & 66.7 & 61.1 & 71.1
& 45.0 & \underline{61.2} & 44.7 & 39.4 & 40.0 & 14.8 & 30.1 & 39.3
& \underline{80.6} & \underline{63.3} \\
Skill-SD
& 81.8 & 43.3 & 56.9 & 39.8 & 42.0 & 45.3 & 51.5
& 44.1 & 60.7 & 43.8 & 39.3 & 40.2 & \underline{16.3} & \underline{34.4} & 39.8
& 75.6 & 58.8 \\
RLSD
& 88.6 & 56.4 & 76.5 & \textbf{75.0} & 60.0 & 58.3 & 69.1
& 44.0 & 59.6 & 43.0 & \textbf{40.3} & \textbf{43.0} & \textbf{18.2} & \underline{34.4} & \underline{40.4}
& 79.7 & 62.4 \\
SDAR
& \underline{96.2} & 61.5 & \textbf{86.4} & \underline{70.8} & \underline{69.3} & \underline{65.3} & \underline{74.9}
& 44.5 & 60.0 & 43.6 & 39.5 & 37.5 & 15.7 & 31.2 & 38.9
& 72.3 & 57.5 \\
AgentOPSD
& 91.4 & 53.8 & 82.7 & 58.3 & 65.3 & 59.7 & 68.6
& 43.0 & 59.4 & \underline{45.1} & 36.7 & 38.3 & 12.9 & 24.8 & 37.2
& 71.3 & 54.7 \\
TASPO
& 94.3 & 53.8 & 77.8 & \textbf{75.0} & 64.0 & \textbf{70.8} & 72.6
& 44.5 & \textbf{62.0} & 44.7 & 38.5 & 38.4 & 15.3 & 31.2 & 39.2
& 74.7 & 62.4 \\
AlignOPSD
& 82.9 & 53.8 & 70.4 & 50.0 & \textbf{80.0} & 58.3 & 65.9
& 44.7 & 60.3 & 44.7 & 39.8 & 40.3 & 14.5 & 28.8 & 39.0
& 75.9 & 58.6 \\
\midrule
\rowcolor{SGRrow}
\textbf{SGR-OPSD}
& \textbf{97.1} & \textbf{69.2} & 81.5 & \textbf{75.0} & 68.0 & \textbf{70.8} & \cellcolor{SGRavg}\textbf{76.9}
& \textbf{45.4} & \textbf{62.0} & \textbf{45.4} & \textbf{40.3} & \underline{41.5} & 15.2 & \textbf{34.6} & \cellcolor{SGRsearch}\textbf{40.6}
& \textbf{85.5} & \cellcolor{SGRsucc}\textbf{68.8} \\
\midrule
\multicolumn{18}{l}{\textit{Qwen2.5-7B-Instruct}} \\
\midrule
Vanilla
& 36.1 & 22.2 & 3.1 & 0.0 & 0.0 & 0.0 & 10.2
& 25.2 & 50.8 & 29.5 & 29.0 & 29.0 & 10.4 & 23.7 & 28.2
& 5.9 & 1.6 \\
GRPO
& 90.5 & 69.2 & \underline{88.9} & 79.2 & 73.3 & 68.1 & 78.2
& \textbf{48.2} & 64.0 & 45.8 & 43.9 & 45.0 & 20.2 & 43.2 & 44.3
& 81.7 & 70.6 \\
Skill-Prompt
& 51.7 & 50.0 & 32.3 & 5.3 & 4.3 & 0.0 & 23.9
& 30.9 & 52.1 & 32.7 & 32.7 & 27.9 & 12.7 & \textbf{66.1} & 36.4
& 1.7 & 0.8 \\
OPSD
& 50.0 & 60.0 & 22.7 & 21.4 & 17.6 & 9.5 & 30.2
& 8.8 & 8.6 & 17.5 & 2.5 & 4.2 & 0.5 & 1.2 & 6.2
& 4.5 & 2.3 \\
GRPO+OPSD
& \textbf{97.1} & 69.2 & \underline{88.9} & 62.5 & 76.0 & 75.0 & 78.1
& 47.2 & 63.8 & 45.6 & 42.7 & 41.8 & 19.5 & 40.8 & 43.1
& 85.2 & 72.8 \\
Skill-SD
& 93.9 & 67.3 & 79.4 & 54.7 & 61.0 & 82.3 & 73.1
& \underline{48.0} & 64.4 & 45.8 & 43.9 & 40.7 & 19.4 & 44.8 & 43.9
& 82.2 & 69.4 \\
RLSD
& 91.4 & 69.2 & 76.5 & 77.1 & 56.0 & 58.3 & 71.4
& 47.2 & 62.8 & \underline{46.1} & \underline{44.4} & 45.8 & 20.6 & 39.2 & 43.7
& \underline{85.6} & 74.1 \\
SDAR
& 94.3 & 71.8 & 84.0 & \textbf{93.8} & 72.0 & 58.3 & 79.0
& 47.9 & 64.4 & \textbf{47.6} & \underline{44.4} & \underline{46.2} & 20.8 & 44.8 & \underline{45.2}
& 84.5 & 72.4 \\
AgentOPSD
& \underline{96.2} & 69.2 & \textbf{91.4} & 83.3 & 72.0 & 69.4 & \underline{80.3}
& 45.5 & 60.4 & 42.5 & 41.7 & 33.3 & 18.6 & 40.8 & 40.4
& 83.0 & 74.0 \\
TASPO
& 88.6 & \textbf{76.9} & 85.2 & 62.5 & 60.0 & \underline{83.3} & 76.1
& 47.5 & \textbf{65.1} & 46.0 & 44.0 & 44.5 & 18.7 & 41.6 & 43.9
& 80.5 & 72.8 \\
AlignOPSD
& 85.7 & 61.5 & 85.2 & \underline{87.5} & \underline{80.0} & 50.0 & 75.0
& \underline{48.0} & 64.5 & \underline{46.1} & \textbf{45.0} & \textbf{47.2} & \underline{21.0} & 42.4 & 44.9
& 85.3 & \underline{74.8} \\
\midrule
\rowcolor{SGRrow}
\textbf{SGR-OPSD}
& \textbf{97.1} & \underline{73.2} & 85.2 & 84.6 & \textbf{89.6} & \textbf{85.4} & \cellcolor{SGRavg}\textbf{85.8}
& 46.8 & \underline{64.7} & 45.9 & 43.3 & 43.9 & \textbf{25.4} & \underline{54.6} & \cellcolor{SGRsearch}\textbf{46.4}
& \textbf{88.6} & \cellcolor{SGRsucc}\textbf{78.2} \\
\bottomrule
\end{tabular}%
}
\caption{Results on ALFWorld, SearchQA, and WebShop (\%); Avg denotes the macro-average. All baseline results are obtained from our own reruns. \textbf{Bold} and \underline{underline} mark the best and second-best results within each backbone.}
\label{tab:main-results}
\end{table*}

\subsection{Subgoal Progress Graph}
\label{sec:graph}
To complement teacher preferences with outcome evidence, we merge matched states across sampled rollouts into a \emph{Subgoal Progress Graph}. Transition counts define empirical values:
\begin{equation}
\begin{aligned}
 \widehat Q(v,a)&=\widehat{\mathbb E}[r+\gamma I\widehat V(v')\mid v,a],\\
 \widehat V(v)&=\sum_a\widehat\pi(a\mid v)\widehat Q(v,a).
\end{aligned}
\label{eq:graph-value}
\end{equation}
Here $v,v'$ are current and successor nodes, $r\in\{0,1\}$ is the success reward, and $\gamma\in(0,1)$ discounts continuation. The expectation averages recorded transitions; $\widehat\pi(a\mid v)$ is the fraction of visits to $v$ taking action $a$. The indicator $I$ permits continuation only for nonterminal transitions with remaining budget and a successor recorded as a source node. The relative graph signal is
\begin{equation}
\begin{aligned}
 \delta_{i,t}&=\widehat Q(v_{i,t},a_{i,t})-\widehat V(v_{i,t}),\\
 g_{i,t}&=\operatorname{clip}(\delta_{i,t},-b_G,b_G).
\end{aligned}
\label{eq:graph-score}
\end{equation}
Here $v_{i,t}$ is the pre-action node, $b_G>0$ bounds the signal, and positive $\delta_{i,t}$ indicates above-average empirical return.

\textbf{Graph-derived action comparisons.}
The Subgoal Progress Graph determines action ordering; calibration translates these preferences into bounded weight adjustments. Beyond node matching for value estimation, comparisons require matching plans, student contexts, and remaining budgets, with all associated visits scoreable and having nonzero advantages:
\begin{equation}
\resizebox{0.86\columnwidth}{!}{$
 \bar g(a)=H_a^\top g,\qquad
 a\succ b\iff\bar g(a)>\bar g(b).
 $}
\label{eq:action-comparison}
\end{equation}
Here $g$ collects turn signals; $H_a$ averages unique visits equally within trajectories, then across trajectories. Preferred pairs $e=(a,b)$ form $\mathcal E$. The same $H_a$ aggregates credit corrections below; graph-score magnitudes do not set the correction margin.

\textbf{Joint credit calibration.}
Graph-preferred actions should receive stronger reinforcement for positive advantages and softer penalties for negative ones. We parameterize the turn weights as
\begin{equation}
 w_{i,t}(c)=1+s_{i,t}c_{i,t}.
\label{eq:correction}
\end{equation}
Here $c$ collects the credit corrections $c_{i,t}$ and $w(c)$ their resulting weights. With $s_{i,t}$ defined above, increasing $c_{i,t}$ favors the action under either advantage sign. We balance teacher retention against graph preferences:
\begin{equation}
\resizebox{0.86\columnwidth}{!}{$
 c^*=\arg\min_c\bigl\{T_GD_p(w(c)\|w^{\mathrm{ref}})
 +\beta\mathcal R_G(c)\bigr\},
 $}
\label{eq:joint}
\end{equation}
where $D_p$ is defined in Equation~\ref{eq:weighted-kl}, $T_G>0$ controls anchoring to the teacher reference $w^{\mathrm{ref}}$, and $\beta\geq0$ scales the graph preference penalty
\begin{equation}
 \mathcal R_G(c)=\frac{1}{2}\sum_{e\in\mathcal E}\rho_e
 [\mu-b_e^\top c]_+^2.
\label{eq:graph-penalty}
\end{equation}
Here $b_e=H_a-H_b$ for each preferred pair $e=(a,b)\in\mathcal E$, so $b_e^\top c$ is the difference in mean corrections. With $[u]_+=\max(u,0)$, the penalty encourages this difference to reach margin $\mu\geq0$. The weight $\rho_e$ normalizes comparison count and coefficient scale (Appendix~\ref{app:implementation}). The optimization is subject to
\begin{equation}
\begin{aligned}
 &\sum_{t=1}^{N_i}p_{i,t}s_{i,t}c_{i,t}=0\quad\forall i,\\
 &|c_{i,t}|\leq\epsilon,\qquad c_{i,t}=0\ \text{if }A_{i,t}=0.
\end{aligned}
\label{eq:credit-constraints}
\end{equation}
These constraints preserve the token-weighted budget and weight bounds, with zero-advantage turns unchanged. Graph comparisons therefore redistribute emphasis within each trajectory; trajectories without eligible comparisons retain their teacher reference exactly.

\subsection{Policy Optimization}
\label{sec:policy}
The calibrated weights redistribute outcome-based credit through
\begin{equation}
 \widetilde A_{i,t}=A_{i,t}w_{i,t}^*.
\label{eq:advantage}
\end{equation}
Here $w_{i,t}^*=w_{i,t}(c^*)$ is the calibrated weight from Equation~\ref{eq:joint}. The resulting advantage $\widetilde A_{i,t}$ is shared by all effective response tokens in the turn, including reasoning tokens. The student minimizes
\begin{equation}
\resizebox{0.86\columnwidth}{!}{$
 \mathcal L_{\mathrm{SGR}}(\theta)=
 \mathcal L_{\mathrm{policy}}\!\left(\theta;\operatorname{sg}(\widetilde A)\right)
 +\lambda_{\mathrm{KL}}\mathcal L_{\mathrm{KL}}(\theta),
 $}
\label{eq:policy}
\end{equation}
Here $\theta$ denotes the student parameters, $\mathcal L_{\mathrm{policy}}$ is the base clipped policy loss, and $\operatorname{sg}$ stops gradients through the advantages. The term $\mathcal L_{\mathrm{KL}}$ regularizes the student toward the reference policy with coefficient $\lambda_{\mathrm{KL}}$; 
Teacher and graph guidance thus enter the policy update through $\widetilde A$.


\section{Experiments}
\label{sec:experiments}
We evaluate \textbf{SGR-OPSD} across three interactive environments and two model scales to examine whether subgoal-guided, graph-calibrated credit improves policy learning.

\subsection{Experimental Setup}
\label{sec:setup}
\textbf{Benchmarks and metrics.}
For ALFWorld~\citep{shridhar2020alfworld}, we evaluate success on the 140-task seen split across six household task categories: Pick, Look, Clean, Heat, Cool, and Pick2. For WebShop~\citep{yao2022webshop}, we evaluate product search and selection on 500 goals, reporting task score and full-task success rate. SearchQA follows Search-R1~\citep{jin2025searchr1}; we report exact-match accuracy on 51,713 questions across NQ, TriviaQA, PopQA, HotpotQA, 2WikiMultiHopQA, MuSiQue, and Bamboogle. Together, these environments test sequential state changes, constraint satisfaction, and evidence gathering.

\textbf{Baselines.}
We compare with Vanilla and Skill-Prompt as training-free references, GRPO~\citep{shao2024deepseekmath} as the outcome-based learner, and OPSD~\citep{zhao2026selfdistilledreasoner}, GRPO+OPSD, Skill-SD~\citep{wang2026skillsd}, RLSD~\citep{yang2026rlsd}, SDAR~\citep{lu2026sdar}, and AgentOPSD~\citep{wang2026agentopsd} as self-distillation baselines. Qwen3.5-Plus and DeepSeek-V3.1 provide additional closed-source model references. 

\textbf{Training configuration.}
We use the Qwen2.5-3B-Instruct and Qwen2.5-7B-Instruct backbones and carry out three independent training runs with seeds $\{0,42,2026\}$. We train \textbf{SGR-OPSD} on 8 NVIDIA A100 GPUs for 150 policy updates, with eight rollouts per task and a learning rate of $10^{-6}$. The maximum interaction lengths are 50, 15, and 4 turns for ALFWorld, WebShop, and SearchQA, respectively. We set $\epsilon=0.1$, $T=T_G=1$, $\beta=2$, $\mu=0.02$, and $\gamma=0.1$. We provide subgoal-generation prompts in Appendix~\ref{app:subgoal-generation} and environment-specific training and evaluation settings in Appendix~\ref{app:experimental-details}.

\subsection{Main Results}
\label{sec:main-results}
\textbf{Stronger multi-turn task completion.}
\textbf{SGR-OPSD} combines task-specific subgoals with the Subgoal Progress Graph to calibrate turn-level credit. As shown in Table~\ref{tab:main-results}, this method improves task completion on both model scales, achieving the highest ALFWorld macro-average and top performance on both WebShop metrics among all baselines. At 7B scale, it achieves \textbf{85.8\%} success on ALFWorld and \textbf{78.2\%} on WebShop. The performance gain over GRPO holds across both scales, with improvements of 7.6 pp. on ALFWorld at both scales and 5.7--7.6 pp. on WebShop success, underscoring the benefit of augmenting outcome feedback with structured credit allocation.

\textbf{Gains beyond existing Self-distillation methods.}
The advantage persists against GRPO+OPSD and the Skill-guided methods Skill-SD, SDAR, and AgentOPSD on both evaluated environments at each model scale. Relative to the strongest baseline for each backbone, \textbf{SGR-OPSD} improves the ALFWorld macro-average by 2.0--5.5 pp and WebShop success by 3.4--5.5 pp. Its leading 7B results on Cool and Pick2 sub-tasks provide concrete examples: these tasks require completing an object-state change or collecting two distinct objects before final placement. The results support the full method's effectiveness on tasks whose success depends on satisfying intermediate requirements.

\textbf{Generalizability to retrieval-based reasoning.}
SearchQA tests the method in a different action space, where progress depends on retrieving evidence and resolving information dependencies. \textbf{SGR-OPSD} improves macro-average accuracy over GRPO by 1.2 and 2.1 points at 3B and 7B, respectively. It achieves the highest macro-average at both scales, exceeding the strongest baseline by 0.2/1.2 pp at 3B/7B, with the best 7B results on MuSiQue. These gains show that the approach benefits retrieval-based reasoning beyond environment manipulation. Across the six headline settings, \textbf{SGR-OPSD}'s across-seed standard deviation ranges from 0.01 to 0.52 pp (Table~\ref{tab:run-variance}).

\subsection{Component Ablations}
\label{sec:ablations}
The ablations examine three components: the Subgoal Progress Graph, teacher evidence, and task-specific subgoal plans (Tables~\ref{tab:ablations} and~\ref{tab:ablations-7b}). The metrics are ALFWorld macro-average seen success, SearchQA macro-average exact-match accuracy, and WebShop success rate.

\begin{table}[t]
\centering
\footnotesize
\setlength{\tabcolsep}{2pt}
\renewcommand{\arraystretch}{1.08}
\begin{tabular*}{\columnwidth}{@{\extracolsep{\fill}}>{\raggedright\arraybackslash}p{0.40\columnwidth}*{3}{c}@{}}
\toprule
\textbf{Method} & ALFWorld & SearchQA & WebShop \\
\midrule
\textbf{SGR-OPSD} & 76.9 & 40.6 & 68.8 \\
\midrule
w/o Graph & 75.4 & 39.5 & 64.8 \\
w/o Teacher Signal & 74.1 & 40.8 & 62.2 \\
Skill instead of Subgoal & 74.6 & 37.7 & 60.4 \\
\bottomrule
\end{tabular*}
\caption{Component ablations with Qwen2.5-3B-Instruct (\%): ALFWorld macro-average seen success, SearchQA macro-average exact-match accuracy, and WebShop success rate.}
\label{tab:ablations}
\end{table}

\begin{table}[t]
\centering
\footnotesize
\setlength{\tabcolsep}{2pt}
\renewcommand{\arraystretch}{1.08}
\begin{tabular*}{\columnwidth}{@{\extracolsep{\fill}}>{\raggedright\arraybackslash}p{0.40\columnwidth}*{3}{c}@{}}
\toprule
\textbf{Method} & ALFWorld & SearchQA & WebShop \\
\midrule
\textbf{SGR-OPSD} & 85.8 & 46.4 & 78.2 \\
\midrule
w/o Graph & 76.7 & 43.8 & 74.2 \\
w/o Teacher Signal & 79.9 & 44.4 & 72.6 \\
Skill instead of Subgoal & 75.9 & 44.2 & 68.4 \\
\bottomrule
\end{tabular*}
\caption{Component ablations with Qwen2.5-7B-Instruct (\%), using the same metrics as Table~\ref{tab:ablations}.}
\label{tab:ablations-7b}
\end{table}

\textbf{Role of graph calibration.}
Relative to \textit{w/o Graph}, which removes the Subgoal Progress Graph and directly uses $w^{\mathrm{ref}}$, the full method scores 4.0 pp. higher on WebShop at both model scales, and 1.1/2.6 pp. higher on SearchQA at 3B/7B. These consistent improvements support calibrating teacher evidence with rollout-derived progress preferences. On ALFWorld, graph calibration improves success by 1.5 pp at 3B and 9.1 pp at 7B, with a larger contribution at the higher model scale.

\begin{table}[t]
\centering
\small
\setlength{\tabcolsep}{0pt}
\renewcommand{\arraystretch}{1.08}
\begin{tabular}{@{}>{\raggedright\arraybackslash}p{\columnwidth}@{}}
\toprule
\textbf{Skill: reusable workflow} \\
\addlinespace[3pt]
Always execute clean tasks in the fixed sequence: \\
(1) locate and acquire target object, \\
(2) bring it to an available water source or sink to clean, \\
(3) navigate to final location, \\
(4) place object. \\
\midrule
\textbf{Subgoal: task-specific requirement} \\
\addlinespace[3pt]
5. Put the clean soapbar on the countertop \\
\textit{Preconditions:} Countertop is found and reachable, soapbar is clean and held \\
\bottomrule
\end{tabular}
\caption{Skill and Subgoal excerpts for the same task, reformatted for readability. Full context is in Appendix~\ref{app:subgoal-generation}.}
\label{tab:guidance-example}
\end{table}

\textbf{Role of teacher evidence.}
Teacher guidance complements preferences from the Subgoal Progress Graph on ALFWorld and WebShop. Compared to \textit{w/o Teacher Signal}, which initializes graph calibration from unit weights, the full method achieves 2.8/5.9 pp. higher success rates on ALFWorld and 6.6/5.6 pp. higher on WebShop at 3B/7B. This pattern underscores the necessity of pairing explicit task requirements with observed progress evidence. SearchQA shows a 2.0 pp. gain at 7B and comparable scores at 3B (40.6\% vs.\ 40.8\%).

\textbf{Choice of privileged guidance.}
Subgoal plans consistently outperform Skill guidance across all available comparisons, achieving gains of 8.4/9.8 pp on WebShop, 2.9/2.2 pp on SearchQA at 3B/7B, and 2.3/9.9 pp on ALFWorld at 3B/7B, under an identical allocation framework. While both representations encode ordered workflows, these results highlight the benefit of instantiating guidance with task-specific objects, destinations, and intermediate state requirements. Appendix~\ref{app:guidance-graph-controls} reports additional comparisons with length-matched Skill and randomized graph preferences.

\subsection{Subgoal Guidance Analysis}
\label{sec:subgoal-guidance-analysis}
\textbf{Task-specific requirements.}
Table~\ref{tab:guidance-example} illustrates this distinction for cleaning a soapbar and placing it on a countertop. The Skill rule specifies a general clean-task workflow; the Subgoal plan names the soapbar and countertop and makes placement conditional on the soapbar being clean and held. The contrast is concrete: the Skill rule refers to a ``target object'' and ``final location,'' while the plan specifies which object must reach which destination and in what state. These explicit completion conditions give teacher scoring a task-specific reference for assessing sampled actions. Through the teacher--student likelihood gap, the plan supplies credit evidence tied to the instruction's intermediate requirements at each decision point.

\textbf{Compact privileged context.}
Across 601 ALFWorld plans, Subgoal guidance averages 87.79 tokens, whereas the six task-type Skill inputs range from 861 to 1,168 tokens. Counts use the Qwen2.5-3B-Instruct tokenizer without special tokens and exclude interaction history. The compact plan retains task-specific goals and preconditions, and is reused unchanged across turns. 
Appendix~\ref{app:subgoal-generation} provides the generation prompts and full comparison.

\begin{figure}[t]
    \centering
    \includegraphics[width=0.45\textwidth]{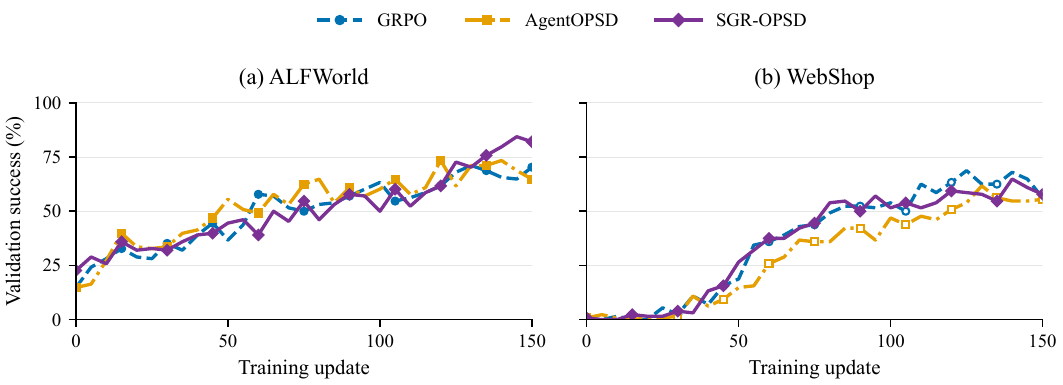}
    \caption{Validation success of GRPO, AgentOPSD, and \textbf{SGR-OPSD} with Qwen2.5-3B-Instruct on ALFWorld and WebShop.}
    \label{fig:training-methods-3b}
\end{figure}

\begin{figure}[t]
    \centering
    \includegraphics[width=0.5\textwidth]{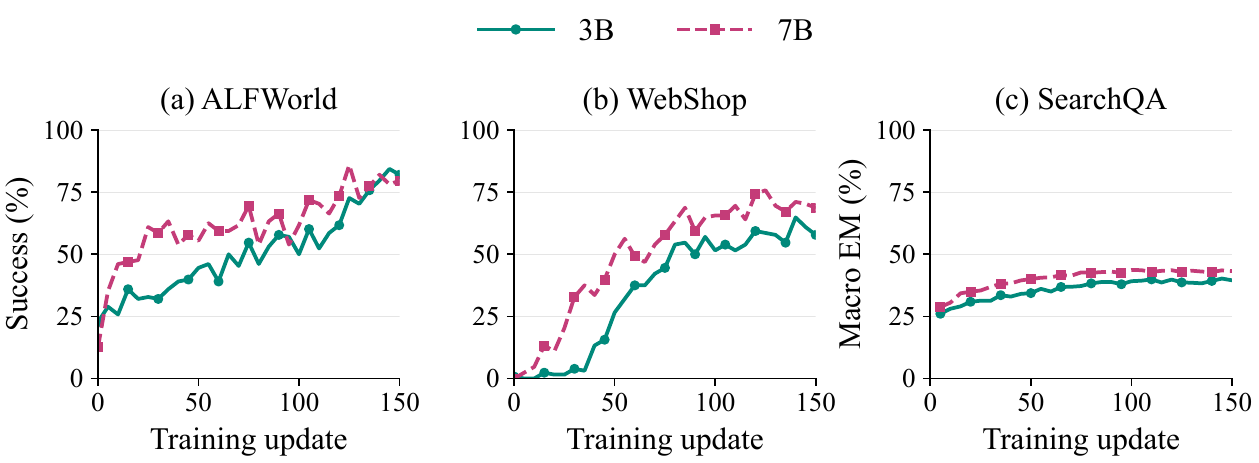}
    \caption{\textbf{SGR-OPSD} with Qwen2.5-3B/7B-Instruct: validation success on ALFWorld/WebShop and macro-average EM on SearchQA.}
    \label{fig:training-sgr-scales}
\end{figure}

\subsection{Training Dynamics}
\label{sec:training-dynamics}
We track validation performance every five training updates, comparing methods at 3B and model scales for \textbf{SGR-OPSD}.

\textbf{Continued gains in later training.}
On ALFWorld, \textbf{SGR-OPSD} sustains its performance lead across the final four validation checkpoints (updates 135--150), finishing 11.7 and 17.2 pp ahead of GRPO and AgentOPSD, respectively (Figure~\ref{fig:training-methods-3b}). While the competing methods overlap during early and intermediate training, \textbf{SGR-OPSD} establishes a sustained late-stage advantage, ultimately reaching 82.0\% success. This trajectory confirms that the final improvement reflects a consistent separation maintained throughout the closing updates. On WebShop, all three methods progress from near-zero success to over 50\%; their final evaluations under the standard protocol are detailed in Table~\ref{tab:main-results}.

\textbf{Learning across scales and environments.}
Figure~\ref{fig:training-sgr-scales} illustrates performance gains across both model scales in all three environments. The 7B model accelerates earlier on ALFWorld and WebShop, whereas the 3B variant exhibits steady late-stage gains. On SearchQA, macro-average Exact Match (EM) accuracy rises from 26.1\% to 39.5\% at 3B scale and from 28.8\% to 43.2\% at 7B scale between updates 5 and 150. The 7B trajectory maintains a clear margin over the 3B curve throughout training on this benchmark, with both scales reaching stability in the closing updates.

\begin{figure}[t]
\centering
\includegraphics[width=0.85\columnwidth]{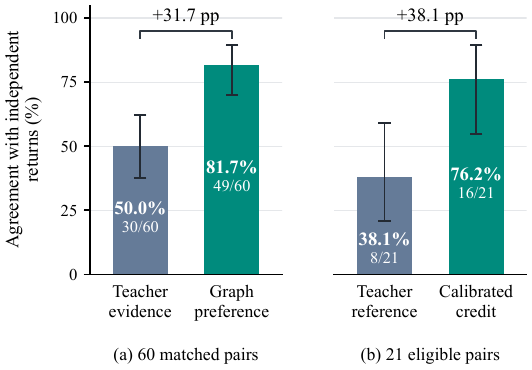}
\caption{Agreement with independent return rankings on ALFWorld (3B): (a) teacher evidence versus Subgoal Progress Graph preferences; (b) credit before and after calibration. Error bars show 95\% Wilson intervals.}
\label{fig:independent-credit-agreement}
\end{figure}

\subsection{Credit Assignment Analysis}
\label{sec:credit-analysis}

\textbf{Independent progress verification.}
Preferences from the Subgoal Progress Graph agree with independent return rankings on 49/60 action pairs (81.7\%), versus 30/60 (50.0\%) for teacher evidence (Figure~\ref{fig:independent-credit-agreement}a). Using a fixed Qwen2.5-3B-Instruct checkpoint, we test one pair per ALFWorld task with 32 continuations per action from matched states, contexts, and budgets. Mean discounted returns determine the reference ranking; the graph and pairs are frozen before sampling these independent continuations. Graph-preferred actions also achieve 2.8 pp higher continuation success.

\textbf{Credit calibration analysis.}
Calibration using Subgoal Progress Graph preferences translates this progress evidence into better credit allocation (Figure~\ref{fig:independent-credit-agreement}b). Capturing the actual paired calibration traces during the training process of ALFWorld, application of Equation~\ref{eq:joint} raises ranking agreement from 8/21 (38.1\%) to 16/21 (76.2\%) on 21 eligible pairs selected independently of continuation outcomes. Sign-aware credit measures support through stronger reinforcement or reduced penalties, depending on the advantage sign. This controlled comparison shows that calibration aligns credit with independently observed returns on these pairs. More details in Appendix~\ref{app:credit-calibration-analysis}.

\textbf{Action-level case study.}
For the tomato-cooling task (Table~\ref{tab:credit-case}), visiting garbagecan~1 reveals a tomato, whereas cabinet~1 is closed. Independent continuations from matched states favor the former in discounted return and success (32/32 vs 13/32). Relative to the teacher reference, calibration strengthens reinforcement for the positive-advantage garbagecan visit and penalties for the negative-advantage cabinet visits, shifting the credit gap from $-0.0109$ to $+0.0191$.

\begin{table}[t]
\centering
\footnotesize
\setlength{\tabcolsep}{2pt}
\begin{tabular*}{\columnwidth}{@{\extracolsep{\fill}}clrrrr@{}}
\toprule
Trajectory & Action & $A_{i,t}$ & Base & $w^{\mathrm{ref}}$ & $w^*$ \\
\midrule
$\tau_1$ & go to garbagecan 1 & $+2.735$ & 1.0000 & 0.9596 & 0.9738 \\
$\tau_2$ & go to cabinet 1 & $-0.365$ & 1.0000 & 1.0397 & 1.0540 \\
$\tau_3$ & go to cabinet 1 & $-0.365$ & 1.0000 & 1.0192 & 1.0366 \\
\bottomrule
\end{tabular*}
\caption{Credit calibration at a matched state in the tomato-cooling task. Each row is a distinct trajectory; Base denotes the unit multiplier on GRPO advantages.}
\label{tab:credit-case}
\end{table}

\subsection{Generalization to Unseen Environments}
\label{sec:unseen-generalization}
\textbf{SGR-OPSD} extends its performance gains to unseen ALFWorld environments, outperforming the strongest baseline by 1.0 pp at 3B and 2.7 pp at 7B (Table~\ref{tab:unseen-generalization}). It achieves \textbf{73.1\%} and \textbf{82.3\%} success on unseen environments at 3B and 7B, respectively, demonstrating transfer capabilities beyond seen settings. 
Subgoal guidance and Subgoal Progress Graph calibration improve policy transfer to unseen environments, where evaluation uses the student's interaction context.
\begin{table}[t]
\centering
\small
\setlength{\tabcolsep}{3pt}
\begin{tabular*}{\columnwidth}{@{\extracolsep{\fill}}lcccc@{}}
\toprule
& \multicolumn{2}{c}{Qwen2.5-3B} & \multicolumn{2}{c}{Qwen2.5-7B} \\
\cmidrule(lr){2-3}\cmidrule(lr){4-5}
Method & Seen & Unseen & Seen & Unseen \\
\midrule
GRPO & 69.3 & 67.5 & 78.2 & 79.6 \\
AgentOPSD & 68.6 & 68.2 & 80.3 & 78.7 \\
AlignOPSD & 65.9 & 72.1 & 75.0 & 65.8 \\
\textbf{SGR-OPSD} & \textbf{76.9} & \textbf{73.1} & \textbf{85.8} & \textbf{82.3} \\
\bottomrule
\end{tabular*}
\caption{ALFWorld seen and unseen macro-average success (\%; means across three training seeds).}
\label{tab:unseen-generalization}
\end{table}

\section{Conclusion}
\label{sec:conclusion}
We introduce \textbf{SGR-OPSD}, which combines subgoal-conditioned teacher evidence with calibration by the Subgoal Progress Graph for turn-level credit assignment across agent interactions, achieving 85.8\% ALFWorld macro-average success and 78.2\% WebShop success at 7B.


\section*{Limitations}
\textbf{Evaluation scope.}
We evaluate \textbf{SGR-OPSD} on three benchmarks using Qwen2.5-3B/7B-Instruct. Its applicability to other model families and open-ended environments requires further evaluation, particularly when observations and action spaces differ from those studied here.

\textbf{Fixed subgoal guidance.}
Subgoal plans are generated from task instructions before interaction and remain fixed across turns. They may omit requirements or fail to accommodate new information revealed by the environment. Updating plans from observed feedback is a natural extension of the current framework.

\textbf{Rollout-dependent graph evidence.}
The Subgoal Progress Graph estimates empirical action values from recorded transitions and outcomes; comparison coverage depends on exploration and the availability of matched action visits. When eligible comparisons are absent, the method retains the teacher reference. Graph scores summarize progress within sampled rollouts rather than establish true action values; broader exploration and more reliable state matching are directions for extending this evidence.

\bibliography{custom}

@misc{wang2026agentopsd,
  author = {Wang, Zi-Han and Lu, Zhengxi and Yao, Zhiyuan and Wu, Jinyang and Wu, Jie and Cai, Zhengzhou and Sun, Yueqing and Ye, Ziang and Hao, Linji and Gu, Qi and Cai, Xunliang and Shen, Yongliang and Yang, Yujiu},
  title = {{AgentOPSD: Recursive Self-Distillation for Agentic Reinforcement Learning}},
  year = {2026},
  howpublished = {arXiv:2608.05987},
  doi = {10.48550/ARXIV.2608.05987},
  url = {https://arxiv.org/abs/2608.05987},
}

@misc{shridhar2020alfworld,
  author = {Shridhar, Mohit and Yuan, Xingdi and C{\^o}t{\'e}, Marc-Alexandre and Bisk, Yonatan and Trischler, Adam and Hausknecht, Matthew},
  title = {{ALFWorld: Aligning Text and Embodied Environments for Interactive Learning}},
  year = {2020},
  howpublished = {arXiv:2010.03768},
  doi = {10.48550/ARXIV.2010.03768},
  url = {https://arxiv.org/abs/2010.03768},
}

@misc{yao2022webshop,
  author = {Yao, Shunyu and Chen, Howard and Yang, John and Narasimhan, Karthik},
  title = {{WebShop: Towards Scalable Real-World Web Interaction with Grounded Language Agents}},
  year = {2022},
  howpublished = {arXiv:2207.01206},
  doi = {10.48550/ARXIV.2207.01206},
  url = {https://arxiv.org/abs/2207.01206},
}

@misc{yang2026taspo,
  author = {Yang, Jingxiao and Gan, Wangjie and Zhuang, Yingxuan and Zhang, Wenqi and Chen, Jintao and Zhang, Xuhong},
  title = {{Reconciling Process Supervision with Outcome-Based Credit in Agentic Policy Optimization}},
  year = {2026},
  howpublished = {arXiv:2608.31077},
  doi = {10.48550/ARXIV.2608.31077},
  url = {https://arxiv.org/abs/2608.31077},
}

@misc{fu2026uniopsd,
  author = {Fu, Zenghuang and Li, Zhaoyang and Ai, Qiuyuan and Han, Xiaofeng and Zheng, Zelong and Wu, Haoyu and Fu, Tianyu and Zhao, Chenxu and Wu, Minghui and He, Guannan and Wang, Changwei},
  title = {{UniOPSD: Unifying Outcome and Hindsight Feedback for Agentic Reinforcement Learning}},
  year = {2026},
  howpublished = {arXiv:2609.34810},
  doi = {10.48550/ARXIV.2609.34810},
  url = {https://arxiv.org/abs/2609.34810},
}

@misc{chen2026alignopsd,
  author = {Chen, Mingju and Lv, Can and Liu, Jinrong and Zhang, Huan and Chang, Heng and Zhou, Shiji},
  title = {{Beyond Timestamps: Decision-Aligned On-Policy Distillation for Long-Horizon Agents}},
  year = {2026},
  howpublished = {arXiv:2609.33391},
  doi = {10.48550/ARXIV.2609.33391},
  url = {https://arxiv.org/abs/2609.33391},
}

@misc{wang2026skillsd,
  author = {Wang, Hao and Wang, Guozhi and Xiao, Han and Zhou, Yufeng and Pan, Yue and Wang, Jichao and Xu, Ke and Wen, Yafei and Ruan, Xiaohu and Chen, Xiaoxin and Qi, Honggang},
  title = {{Skill-SD: Skill-Conditioned Self-Distillation for Multi-turn LLM Agents}},
  year = {2026},
  howpublished = {arXiv:2604.10674},
  doi = {10.48550/ARXIV.2604.10674},
  url = {https://arxiv.org/abs/2604.10674},
}

@misc{shao2024deepseekmath,
  author = {Shao, Zhihong and Wang, Peiyi and Zhu, Qihao and Xu, Runxin and Song, Junxiao and Bi, Xiao and Zhang, Haowei and Zhang, Mingchuan and Li, Y. K. and Wu, Y. and Guo, Daya},
  title = {{DeepSeekMath: Pushing the Limits of Mathematical Reasoning in Open Language Models}},
  year = {2024},
  howpublished = {arXiv:2402.03300},
  doi = {10.48550/ARXIV.2402.03300},
  url = {https://arxiv.org/abs/2402.03300},
}

@misc{yang2026rlsd,
  author = {Yang, Chenxu and Qin, Chuanyu and Si, Qingyi and Chen, Minghui and Gu, Naibin and Yao, Dingyu and Lin, Zheng and Wang, Weiping and Wang, Jiaqi and Duan, Nan},
  title = {{Self-Distilled RLVR}},
  year = {2026},
  howpublished = {arXiv:2604.03128},
  doi = {10.48550/ARXIV.2604.03128},
  url = {https://arxiv.org/abs/2604.03128},
}

@misc{lu2026sdar,
  author = {Lu, Zhengxi and Yao, Zhiyuan and Han, Zhuowen and Wang, Zi-Han and Wu, Jinyang and Gu, Qi and Cai, Xunliang and Lu, Weiming and Xiao, Jun and Zhuang, Yueting and Shen, Yongliang},
  title = {{Self-Distilled Agentic Reinforcement Learning}},
  year = {2026},
  howpublished = {arXiv:2605.15155},
  doi = {10.48550/ARXIV.2605.15155},
  url = {https://arxiv.org/abs/2605.15155},
}

@article{lv2026pcsd,
  title={Pcsd: Persistent consistency for self-distillation in agentic reinforcement learning},
  author={Lv, Chunji and Wei, Yangguang and Liu, Junlin and Gao, Yang and Liu, Ming and Wang, Xinming and Wu, Jinyang and Wang, Guoren and Li, Changsheng},
  journal={arXiv preprint arXiv:2608.01837},
  year={2026}
}

@article{lan2026trust,
  title={Trust Is Not Enough: Influence Calibration for On-Policy Self-Distillation in Agentic RL},
  author={Lan, Qizhen and Xiao, Xi and Guan, Xiangchen and Fan, Mengchen and Lin, Moule and Choi, Jung Im and Zhu, Lijing},
  journal={arXiv preprint arXiv:2608.14945},
  year={2026}
}

@misc{deepseekai2024deepseekv3technicalreport,
      title={DeepSeek-V3 Technical Report}, 
      author={DeepSeek-AI},
      year={2024},
      eprint={2412.19437},
      archivePrefix={arXiv},
      primaryClass={cs.CL},
      url={https://arxiv.org/abs/2412.19437}, 
}

@misc{zhao2026selfdistilledreasoner,
  author = {Zhao, Siyan and Xie, Zhihui and Liu, Mengchen and Huang, Jing and Pang, Guan and Chen, Feiyu and Grover, Aditya},
  title = {{Self-Distilled Reasoner: On-Policy Self-Distillation for Large Language Models}},
  year = {2026},
  eprint = {2601.18734},
  archivePrefix = {arXiv},
  primaryClass = {cs.LG},
  url = {https://arxiv.org/abs/2601.18734}
}

@misc{jin2025searchr1,
  author = {Jin, Bowen and Zeng, Hansi and Yue, Zhenrui and Yoon, Jinsung and Arik, Sercan and Wang, Dong and Zamani, Hamed and Han, Jiawei},
  title = {{Search-R1: Training LLMs to Reason and Leverage Search Engines with Reinforcement Learning}},
  year = {2025},
  howpublished = {arXiv:2503.09516},
  doi = {10.48550/arXiv.2503.09516},
  url = {https://arxiv.org/abs/2503.09516}
}

\appendix
\label{sec:appendix}
\section{Privileged Information and Subgoal Generation}
\label{app:subgoal-generation}

\subsection{Comparison of Privileged Information}
Table~\ref{tab:privileged-sources} compares task specificity, subgoal planning, and interaction requirements of privileged guidance in related methods. Subgoal plans condition guidance on the current instruction and explicitly connect intermediate goals through ordering preconditions. They are generated before student interaction and reused across turns, making the task's object, destination, count, and required state explicit without requiring a successful rollout or a reference answer.

\begin{table*}[t]
\centering
\small
\newcommand{\PIyes}{\textcolor[RGB]{27,128,70}{$\checkmark$}}
\newcommand{\PIno}{\textcolor[RGB]{190,45,45}{$\times$}}
\setlength{\tabcolsep}{4pt}
\renewcommand{\arraystretch}{1.15}
\begin{tabular*}{\textwidth}{@{}l@{\extracolsep{\fill}}cccc@{}}
\toprule
\textbf{Method}
& \shortstack{\textbf{Task-instance}\\\textbf{specific}}
& \shortstack{\textbf{Task-specific}\\\textbf{subgoal plan}}
& \shortstack{\textbf{No successful}\\\textbf{rollout required}}
& \shortstack{\textbf{No task-specific}\\\textbf{interaction required}} \\
\midrule
OPSD (Skill)~\citep{zhao2026selfdistilledreasoner} & \PIno & \PIno & \PIyes & \PIyes \\
RLSD~\citep{yang2026rlsd} & \PIyes & \PIno & \PIyes & \PIyes \\
Skill-SD~\citep{wang2026skillsd} & \PIyes & \PIyes & \PIyes & \PIno \\
SDAR~\citep{lu2026sdar} & \PIno & \PIno & \PIyes & \PIyes \\
AgentOPSD~\citep{wang2026agentopsd} & \PIno & \PIno & \PIyes & \PIyes \\
UniOPSD~\citep{fu2026uniopsd} & \PIyes & \PIno & \PIno & \PIno \\
\midrule
\textbf{SGR-OPSD} & \PIyes & \PIyes & \PIyes & \PIyes \\
\bottomrule
\end{tabular*}
\caption{Comparison of privileged guidance (\PIyes: present; \PIno: absent): task specificity refers to the current instance, and subgoal plans specify intermediate goals with explicit ordering or dependencies. Interaction requirements concern constructing privileged guidance from current-task executions, not ordinary student rollouts.}
\label{tab:privileged-sources}
\end{table*}

\paragraph{Concrete Skill--Subgoal comparison.}
Table~\ref{tab:skill-subgoal-example} pairs a recorded Subgoal plan with the Skill text selected for the same ALFWorld instruction. In this Skill configuration, task-type matching concatenates general rules with the corresponding task-specific file. Both forms encode ordering: Skill expresses reusable rules with generic objects and destinations, while Subgoal instantiates these requirements for the current task. The table reproduces two Skill excerpts and the complete stored plan.

\begin{table*}[t]
\centering
\small
\setlength{\tabcolsep}{6pt}
\begin{tabular}{@{}p{0.47\textwidth}p{0.49\textwidth}@{}}
\toprule
\multicolumn{2}{@{}l}{\textbf{Task:} \textit{clean some soapbar and put it in countertop.}} \\
\midrule
\textbf{Skill guidance: verbatim excerpts} & \textbf{Subgoal guidance: complete recorded plan} \\
\midrule
10. Progressive Goal Decomposition: Break tasks into ordered sub-goals: (1) locate object, (2) obtain or transform it, (3) locate destination or reference object, (4) satisfy spatial relation or placement. Complete each sub-goal before moving to the next. Apply this at the start of every episode.

1. Phase-Ordered Plan: Always execute clean tasks in the fixed sequence: (1) locate and acquire target object, (2) bring it to an available water source or sink to clean, (3) navigate to final location, (4) place object. Apply this as soon as the goal specifies the object must be clean before placement.
&
1. Find a soapbar in the environment Preconditions: None

2. Pick up the soapbar Preconditions: Soapbar is found and reachable

3. Clean the soapbar (e.g., rinse or wash it) so it is clean Preconditions: Soapbar is held

4. Find a countertop receptacle Preconditions: Soapbar is clean and held

5. Put the clean soapbar on the countertop Preconditions: Countertop is found and reachable, soapbar is clean and held \\
\bottomrule
\end{tabular}
\caption{Actual ALFWorld guidance for the same task. Skill uses type-level rules; the recorded Subgoal plan names the soapbar and countertop and ties placement to the object's required state.}
\label{tab:skill-subgoal-example}
\end{table*}

\paragraph{Guidance length.}
We compare the six task-type Skill inputs with an ALFWorld corpus of 601 unique cached Subgoal plans, using the Qwen2.5-3B-Instruct tokenizer without special tokens. Skill inputs ranged from 861 to 1168 tokens; Subgoal plans averaged 87.79 tokens (median 85; range 28--162). These counts cover the privileged text supplied to the teacher, excluding interaction history. They quantify the compactness of the task-specific representation in this corpus.

\subsection{DeepSeek-V3.1 Generation Prompts}
We query DeepSeek-V3.1~\citep{deepseekai2024deepseekv3technicalreport} through a chat-completion interface with temperature 0 and a maximum output of 1024 tokens. Each initial request contains the environment-specific system prompt reproduced below and a user message containing only the task instruction $x$ (the question text for Search-QA). Outputs that fail validation are retried with the previous response and validation feedback. The deployment model identifier is \texttt{deepseek-v31-meituan}. Plans are cached by task and generation configuration and reused unchanged; ALFWorld and WebShop generate them before the first rollout for a new instruction, while Search-QA precomputes the scheduled training questions. The external model generates guidance, and the rollout policy supplies both likelihood scores.

The response is a JSON object containing 2--8 subgoals, each with nonempty \texttt{goal} and \texttt{precondition} fields. The teacher receives the numbered rendering \texttt{j. <goal> Preconditions: <precondition>}. The complete plan is supplied at each turn, without stage-dependent filtering, and is absent from the student's acting context and evaluation prompts.

\paragraph{ALFWorld system prompt.}
\begin{quote}
\footnotesize\ttfamily\raggedright
Plan an ALFWorld household task from its task description only. Return JSON only: \{"subgoals": [\{"goal": "...", "precondition": "..."\}]\}. Give 2 to 8 concise ordered subgoals, including any necessary exploration, acquisition, cleaning/heating/cooling, and final placement/use. State ordering preconditions. Describe desired states rather than a guessed action transcript. You do not know the initial scene or object locations. Never invent locations, object instance numbers, observations, or completed steps. Use conditional search where needed. Preserve the task's exact object type, destination type, count, and required state. Never substitute another receptacle or add an alternative destination (a table is not interchangeable with a counter). A two-object task requires two distinct objects. Keep the whole plan below 350 English words. Do not include an explanation outside JSON.
\end{quote}

\paragraph{WebShop system prompt.}
\begin{quote}
\footnotesize\ttfamily\raggedright
Plan a WebShop shopping task using only the customer's instruction. Return JSON only: \{"subgoals": [\{"goal": "...", "precondition": "..."\}]\}. Give 2 to 8 concise ordered subgoals and their ordering preconditions. Include searching for candidates, inspecting product details, checking ALL requested attributes, selecting the exact required options, verifying the price constraint, and buying only after the visible evidence supports the requirements. Describe desired states, not a guessed action transcript. Preserve every stated product type, attribute, option (such as size/color), quantity, and price bound. Do not silently relax or substitute a requirement, even if it appears unusual. You have no catalog, search results, product IDs, prices, observations, past actions, or reward information. Never invent these, assert that a suitable product exists, or claim a step is complete. Use conditional search/refinement where needed. Do not add unstated purchase requirements. Keep the plan under 350 English words.
\end{quote}

\paragraph{Search-QA system prompt.}
\begin{quote}
\footnotesize\ttfamily\raggedright
Plan how to answer a retrieval question using only the question. Return JSON only: \{"subgoals": [\{"goal": "...", "precondition": "..."\}]\}. Give 2 to 8 concise ordered evidence goals and their dependencies, under 350 English words. Identify missing facts, bridge entities and constraints that must be established by retrieved evidence. For a single-hop question use a simple evidence lookup and answer verification. For a multi-hop question order lookups according to their dependencies. End with checking the evidence against the question and producing a concise answer. You have no search results, gold answer, supporting-fact labels, student history or reward. Do not supply or guess the answer, invent bridge entities, assert an unknown fact, or claim that a goal is complete. Use placeholders or conditional goals for unknown entities. Preserve the exact entities, relation and time constraints in the question. Do not prescribe more searches than the four-turn interaction budget.
\end{quote}

\section{Implementation Details of Credit Allocation}
\label{app:implementation}
This appendix specifies the scoring, weighting, and graph conventions used by the allocation in Section~\ref{sec:method}.

\subsection{Scoring Context and Action Tokens}
The teacher augments the student's interaction context by prepending the cached subgoal plan. The resulting teacher prompt is left-truncated when it exceeds the environment-specific context limit. Both branches use the same pre-update actor and score the original sampled response IDs. Action-token probabilities remain conditioned on earlier reasoning tokens; Equation~\ref{eq:gap} is not a probability marginalized over reasoning.

The action mask selects original response tokens that overlap the trimmed executable-action span, including tokens crossing its byte boundaries. The span is not retokenized. A turn without a scoreable action receives $d_{i,t}=0$ and is excluded from graph comparisons. Its outcome penalty and full-response policy update remain active. Such a turn still participates in its trajectory's allocation budget, so its multiplier need not remain one.

\subsection{Training Mass and Unique Visits}
Training batches can contain padding copies of an environment turn. The allocator represents each unique turn once, after checking agreement of its copies' evidence and base advantages. Its mass $m_{i,t}$ is the number of effective response tokens summed across all training copies. These copies contribute to the policy loss and hence to $p_{i,t}$, but do not create additional graph visits. The allocated advantage is broadcast back to all copies.

Base advantages retain the reward normalization and invalid-action penalties of the training recipe in Section~\ref{sec:setup}. The allocator does not replace them by a common trajectory advantage. Accordingly, positive, negative, and zero advantages can occur within one trajectory. The multiplier budget $\sum_{t=1}^{N_i}p_{i,t}w_{i,t}=1$ does not imply conservation of the sum of advantages or the optimizer's gradient norm.

\subsection{Graph Construction and Pair Eligibility}
\textbf{State and action identities.}
Each task's group of eight student rollouts forms a separate Subgoal Progress Graph at each update. Recorded visits contribute transition counts once, excluding training padding copies. ALFWorld node keys combine observation-memory anchors with physical-state signatures hashing simulator facts and entity metadata. WebShop keys combine formatted observations with product identity, page, page name, and selected options. These descriptors support graph estimation without entering teacher or student prompts. Within a node, actions are grouped by their exact executed strings.

\textbf{SearchQA states and actions.}
The graph ID hashes the update, rollout-group ID, and question ID. A node represents the exact formatted student context: the question, interaction count, and ordered executed queries with their returned evidence. The four-turn history window covers preceding interactions within the four-turn budget. Nodes merge only when their context strings match within the graph. Different query histories remain separate even if they retrieve the same passage. The pre-action state signature hashes the question ID and context; the student-prompt hash checks the scoring input after prompt processing.

Actions use the projected executable string: \texttt{<search>query</search>} or \texttt{<answer>answer</answer>}. Projection removes surrounding reasoning, standardizes the wrapper, and trims payload-boundary whitespace without semantically normalizing queries or answers. Gold answers and reward labels are excluded from node keys and state signatures.

\textbf{Empirical graph values.}
Let $N(v,a)$ count recorded visits taking action $a$ at node $v$, and $N(v)=\sum_a N(v,a)$. Equation~\ref{eq:graph-value} uses $\widehat\pi(a\mid v)=N(v,a)/N(v)$: $\widehat Q$ averages immediate rewards and discounted successor values over recorded transitions, and $\widehat V$ averages action values by visit frequency. Their difference measures an action's empirical return relative to the state average.

For ALFWorld and WebShop, the recorded binary success reward (0 or 10) is divided by 10 before value estimation; SearchQA uses binary terminal exact-match reward directly. Thus all three environments use $r\in\{0,1\}$ and $\gamma=0.1$, with success assigned to the terminating transition. Terminal transitions, exhausted budgets, and successors absent from the recorded source-node set have zero continuation value; immediate rewards are retained. Failed or budget-truncated episodes remain in the transition counts, with zero terminal success reward.

Values are initialized to zero and iterated until the maximum absolute change is below $10^{-12}$, with at most 200 iterations; failure to converge raises an error. Cycles are handled by these discounted fixed-point updates. Clipping is applied after subtracting $\widehat V$ from $\widehat Q$, as in Equation~\ref{eq:graph-score}; there is no subsequent reward rescaling. All visits to a state--action pair share its empirical signal. The values summarize continuation behavior pooled by node; the additional context and budget checks below restrict comparisons, not the value-estimation pool.

\textbf{Pair eligibility and averaging.}
Within each node--plan bucket, every visit associated with the two candidate actions must be scoreable, have nonzero base advantage, and agree on the pre-action state signature, student-prompt hash, and remaining step budget. ALFWorld signatures use simulator facts and entity metadata; WebShop signatures use recorded session state and available actions. Success reachability is not an additional eligibility condition. We retain $(a,b)$ when $\bar g(a)>\bar g(b)$, excluding ties. The vector $H_a$ averages unique visits equally within each trajectory, then equally across trajectories. Although identical state--action visits share a graph score, this averaging also defines their mean credit correction. Graph scores supply pair ordering, while $\mu$ sets the correction margin. For each eligible pair $e$, the comparison weight is
\begin{equation}
 \rho_e=\frac{1}{|\mathcal E_v|\,\lVert b_e\rVert_2^2},
\label{eq:rho}
\end{equation}
where $\mathcal E_v$ contains the eligible pairs in the same node--plan bucket. This weight normalizes comparison count and coefficient scale; it is not a bootstrap reliability estimate.

\subsection{Optimization and Fallback}
For a nonzero advantage, differentiating the teacher objective before clipping gives $s_{i,t}d_{i,t}-T\log w_{i,t}-\nu_i=0$. The resulting exponential tilt is clipped to the prescribed interval, and the trajectory multiplier $\nu_i$ is found by bisection to satisfy the budget. Zero-advantage turns retain $w_{i,t}=1$. This solves the bounded allocation directly, rather than clipping weights and subsequently renormalizing them outside their bounds.

\textbf{Joint solver.}
We use float64 Nesterov-accelerated projected gradient descent, initialized at the teacher correction $c^{\mathrm{ref}}$, with monotone restart. Let $F(c)$ denote the joint objective, $D=\operatorname{diag}(p)$ the token-mass metric, and $\Pi_D$ its projection onto the bounded, per-trajectory budget constraints. The step size is $1/L$, with
\begin{equation}
 L=\frac{T_G}{1-\epsilon}
   +\beta\bigl\|M^\top\operatorname{diag}(\rho)M\bigr\|_\infty,
\end{equation}
where $B$ has rows $b_e^\top$, $M=|B|D^{-1/2}$, and the matrix infinity norm is the maximum row sum. The momentum coefficient is $(\sqrt L-\sqrt m)/(\sqrt L+\sqrt m)$ with $m=T_G/(1+\epsilon)$. An extrapolated step that increases the objective beyond $10^{-13}(1+|F(c)|)$ is replaced by the ordinary projected-gradient step and momentum is reset. Outside the feasible box, extrapolation uses the KL term's quadratic extension matching its value, gradient, and curvature at the clipped boundary.

We stop when
\begin{equation}
 L\bigl\|c-\Pi_D(c-D^{-1}\nabla F(c)/L)\bigr\|_\infty
 \leq 10^{-8},
\end{equation}
with at most 5,000 iterations. Teacher-budget bisection uses 100 iterations; each metric projection uses 55. A solution must be finite, have maximum trajectory-budget error at most $10^{-10}$ and bound violation at most $10^{-12}$, and increase the objective over the teacher initialization by at most $10^{-9}$. Failure of convergence or these checks raises an error and stops the update; it does not silently substitute teacher weights.

\textbf{No-comparison case.} A trajectory with no eligible graph term retains its teacher reference exactly. Within a trajectory containing graph comparisons, the budget can also adjust turns that do not participate in a pair. The soft hinge term encourages pairwise margins but does not guarantee that every graph comparison is satisfied. The experimental allocation hyperparameters are reported in Section~\ref{sec:setup}.

\section{Experimental Details}
\label{app:experimental-details}
\subsection{Training and Evaluation Protocols}
Table~\ref{tab:config} summarizes the shared rollout and base policy-optimization settings for the trained methods, with environment-specific values shown separately. Method-specific supervision and credit-allocation objectives follow their respective formulations; \textbf{SGR-OPSD}'s allocation parameters are given in Section~\ref{sec:setup}. Vanilla, Skill-Prompt, and the API-based references require no policy training. SearchQA training uses NQ and HotpotQA, with evaluation on all seven datasets.

\begin{table*}[t]
\centering
\small
\setlength{\tabcolsep}{6pt}
\renewcommand{\arraystretch}{1.08}
\begin{tabular*}{\textwidth}{@{\extracolsep{\fill}}lccc@{}}
\toprule
\textbf{Setting} & \textbf{ALFWorld} & \textbf{WebShop} & \textbf{SearchQA} \\
\midrule
Tasks per update & 16 & 16 & 128 \\
Maximum interaction turns & 50 & 15 & 4 \\
Maximum prompt tokens & 2048 & 4096 & 4096 \\
Interaction history (turns) & 2 & 2 & 4 \\
Reference-policy KL coefficient & 0.01 & 0.01 & 0.001 \\
Learning-rate warmup ratio & 0 & 0 & 0.1 \\
PPO minibatch size & 256 & 64 & 256 \\
\midrule
Optimizer & \multicolumn{3}{c}{AdamW} \\
Learning rate & \multicolumn{3}{c}{$10^{-6}$} \\
Optimizer $(\beta_1,\beta_2)$ & \multicolumn{3}{c}{$(0.9,0.999)$} \\
Weight decay / gradient clipping & \multicolumn{3}{c}{0.01 / 1.0} \\
PPO clipping / dual-clip coefficient & \multicolumn{3}{c}{0.2 / 3.0} \\
Policy epochs per batch & \multicolumn{3}{c}{1} \\
Entropy coefficient & \multicolumn{3}{c}{0.001} \\
KL estimator & \multicolumn{3}{c}{Low-variance KL} \\
Rollouts per task / policy updates & \multicolumn{3}{c}{8 / 150} \\
Rollout temperature / top-$p$ & \multicolumn{3}{c}{1.0 / 1.0} \\
Maximum response tokens & \multicolumn{3}{c}{512} \\
Training seeds & \multicolumn{3}{c}{$0,42,2026$} \\
\bottomrule
\end{tabular*}
\caption{Policy optimization and rollout settings. Values shared across environments span all three columns.}
\label{tab:config}
\label{tab:comparison-configs}
\end{table*}

For \textbf{SGR-OPSD}, the reference execution configurations use actor microbatches of 32/8/16 per GPU and rollout tensor parallelism of 2/2/1 for ALFWorld/WebShop/SearchQA, respectively. WebShop uses two GPUs at 3B and eight at 7B. Periodic checkpoints are saved every 10 updates; validation and checkpoint selection are specified below.

\paragraph{Base advantages.}
For ALFWorld and WebShop, row-level rewards are centered and scaled by the mean and sample standard deviation over training rows sharing a task UID, including padding copies, with stabilizer $10^{-6}$. The invalid-action penalty coefficient is $0.1$, so advantages can differ across turns of one trajectory. The policy loss averages over effective response tokens. Allocation deduplicates turns while retaining their training token mass, as described in Appendix~\ref{app:implementation}.

\paragraph{Evaluation sets and decoding.}
Full ALFWorld evaluation uses 140 seen and 134 unseen tasks, with one rollout per task, temperature 0.4, top-$p=1$, and top-$k=-1$. Table~\ref{tab:main-results} uses the seen split and reports the unweighted mean of its six category success rates. WebShop evaluation uses the 500 goals indexed 0--499 with temperature 0.4. Success is the fraction of fully completed goals; Score is the mean native partial-completion reward multiplied by 100. SearchQA uses greedy decoding and question-level exact match across 51,713 questions in seven datasets.

\paragraph{Checkpoint selection.}
For \textbf{SGR-OPSD} on ALFWorld and WebShop, validation tasks are newly sampled from the training task pool every 5 policy updates. We select the saved checkpoint with the highest validation success rate and evaluate it on the fixed evaluation sets described above, which are not used for checkpoint selection. For SearchQA main results and ablations, we evaluate the fixed update-150 checkpoint on all 51,713 questions.

\paragraph{Evaluation randomness.}
The frozen ALFWorld full-evaluation protocol uses environment seed 0, task seed $1000+\text{index}$, and engine seed 0. Training and evaluation seeds are recorded separately; a shared engine seed does not imply paired per-request sampling.
\paragraph{Aggregation.}
Table~\ref{tab:main-results} reports three-seed aggregates for trained methods, using seeds $\{0,42,2026\}$. ALFWorld and SearchQA Avg first average equally over the six task categories and seven datasets, respectively, within each run, then across runs. WebShop Score and success use the same seed aggregation. In Tables~\ref{tab:ablations} and~\ref{tab:ablations-7b}, full-method rows report three-seed means, while ablation variants use seed~0. Figures~\ref{fig:training-methods-3b} and~\ref{fig:training-sgr-scales} show individual seed-0 runs.

\paragraph{Run-to-run variation.}
Table~\ref{tab:run-variance} supplements the main-table means with across-seed standard deviations for \textbf{SGR-OPSD} and six baselines. \textbf{SGR-OPSD} achieves the lowest standard deviation on ALFWorld and SearchQA at both model scales.

\begin{table*}[t]
\centering
\small
\setlength{\tabcolsep}{4pt}
\renewcommand{\arraystretch}{1.08}
\begin{tabular*}{\textwidth}{@{\extracolsep{\fill}}l*{6}{c}@{}}
\toprule
& \multicolumn{3}{c}{Qwen2.5-3B-Instruct} & \multicolumn{3}{c}{Qwen2.5-7B-Instruct} \\
\cmidrule(lr){2-4}\cmidrule(lr){5-7}
Method & ALFWorld & SearchQA & WebShop & ALFWorld & SearchQA & WebShop \\
\midrule
GRPO & $69.3 \pm 9.39$ & $39.4 \pm 3.47$ & $63.1 \pm 2.32$ & $78.2 \pm 5.31$ & $44.3 \pm 4.85$ & $70.6 \pm 4.78$ \\
RLSD & $69.1 \pm 6.07$ & $40.4 \pm 5.37$ & $62.4 \pm 1.06$ & $71.4 \pm 5.47$ & $43.7 \pm 5.36$  & $74.1 \pm 3.80$ \\
SDAR & $74.9 \pm 4.13$ & $38.9 \pm 4.85$ & $57.5 \pm 3.18$ & $79.0 \pm 4.23$ & $45.2 \pm 3.29$ & $72.4 \pm 2.16$ \\
AgentOPSD & $68.6 \pm 4.04$ & $37.2 \pm 2.41$ & $54.7 \pm 0.95$ & $80.3 \pm 4.93$ & $40.4 \pm 3.75$ & $74.0 \pm 4.51$ \\
TASPO & $72.6 \pm 3.85$ & $39.2 \pm 3.74$ & $62.4 \pm 3.34$ & $76.1 \pm 3.64$ & $43.9 \pm 2.54$ & $72.8 \pm 3.25$ \\
AlignOPSD & $65.9 \pm 5.26$ & $39.0 \pm 3.83$ & $58.6 \pm 1.47$ & $75.0 \pm 2.76$ & $44.9 \pm 4.23$ & $74.8\pm 3.22$ \\
\midrule
\textbf{SGR-OPSD} & $76.9 \pm 1.52$ & $40.6 \pm 1.21$ & $68.8 \pm 2.25$ & $85.8 \pm 2.46$ & $46.4 \pm 2.14$ & $78.2 \pm 2.36$ \\
\bottomrule
\end{tabular*}
\caption{Mean $\pm$ standard deviation over three training seeds ($0$, $42$, $2026$): ALFWorld seen macro-average success, SearchQA macro EM, and WebShop success (\%).}
\label{tab:run-variance}
\end{table*}

\subsection{Task-Specific Guidance and Graph Preferences}
\label{app:guidance-graph-controls}

\begin{table}[t]
\centering
\small
\setlength{\tabcolsep}{4pt}
\begin{tabular*}{\columnwidth}{@{\extracolsep{\fill}}lcc@{}}
\toprule
Configuration & Seen & Unseen \\
\midrule
\textbf{SGR-OPSD} & \textbf{76.9} & \textbf{73.1} \\
Length-matched Skill & 71.0 & 68.4 \\
Randomized graph preferences & 72.2 & 63.9 \\
\bottomrule
\end{tabular*}
\caption{ALFWorld macro-average success (\%) with Qwen2.5-3B-Instruct.}
\label{tab:guidance-graph-controls}
\end{table}

\textbf{Task-specific guidance.}
\textbf{SGR-OPSD} outperforms length-matched Skill by 5.9 and 4.7 percentage points on seen and unseen environments, respectively (Table~\ref{tab:guidance-graph-controls}). This control replaces Subgoal plans with complete task-type Skill rules matched within 10\% of their token length, while retaining the same interaction history for teacher scoring. The results highlight the value of guidance tailored to the current task.

\textbf{Directional graph evidence.}
\textbf{SGR-OPSD} outperforms randomized graph preferences by 4.7 and 9.2 percentage points on seen and unseen environments, respectively. This control randomly reverses each eligible action preference with probability 0.5 while preserving the pair structure and calibration constraints. The comparison highlights the importance of informative graph preferences for credit allocation, with the largest performance gap on unseen environments.

\subsection{Training-Curve Sources}
\label{app:curve-protocol}
\textbf{SearchQA validation runs.}
SearchQA curves report greedy-decoding performance on a fixed 3,125-question validation mixture every five updates. Figure~\ref{fig:training-sgr-scales}(c) uses dedicated 3B and 7B runs, separate from the update-150 main-result runs. The mixture contains 500 each from NQ, TriviaQA, PopQA, HotpotQA, 2WikiMultiHopQA, and MuSiQue, and all 125 Bamboogle questions. Macro EM is $100\times\frac{1}{7}\sum_{d=1}^{7} C_d/N_d$, where $C_d$ and $N_d$ are each dataset's correct-answer and question counts. The 60 plotted points are computed from these per-dataset validation counts. The original and curve-run packages use byte-identical graph construction, pair eligibility, and allocation solver files.

The full evaluation pool contains 3,610 NQ, 11,313 TriviaQA, 14,267 PopQA, 7,405 HotpotQA, 12,576 2WikiMultiHopQA, 2,417 MuSiQue, and 125 Bamboogle questions. Main-table and ablation SearchQA scores use this full pool. Both evaluations report macro EM across seven datasets: curves use the fixed subset at each checkpoint, while tables use full-pool checkpoint evaluations.

\textbf{ALFWorld and WebShop curves.}
The other panels use recorded training-validation points from the seed-0 method-comparison and full-method runs, including their documented checkpoint-resume chains. Validation uses 128 episodes per check, and the plotted success is the fraction of successful episodes. ALFWorld curves therefore differ from the six-category macro-average in the result tables. In the 3B WebShop method comparison, GRPO and AgentOPSD use fixed validation tasks while \textbf{SGR-OPSD} samples tasks; final WebShop table entries use all 500 goals. All curves use recorded checkpoints and retain their validation protocols.

\section{Progress Verification and Credit Calibration}
\label{app:progress-credit}

\subsection{Independent Progress Verification}
\label{app:independent-progress}
\textbf{Agreement with independent outcomes.}
Preferences from the Subgoal Progress Graph agree with independent continuation returns on 49/60 matched ALFWorld action pairs (81.7\%), compared with 30/60 (50.0\%) for teacher evidence. Graph-preferred actions also achieve higher mean discounted return and continuation success (Table~\ref{tab:independent-progress-detail}). They succeed in 1,756/1,920 continuations, compared with 1,702/1,920 for their alternatives. At the task level, success is higher for 22 pairs, lower for nine, and tied for 29. On the 31 pairs with unequal success rates, graph preferences agree with the success ordering on 22 pairs (71.0\%), compared with 16 (51.6\%) for teacher evidence. Return agreement evaluates all 60 pairs and captures both completion and its timing.

\textbf{Independent continuation protocol.}
The evaluation fixes Qwen2.5-3B-Instruct at update 150, selected by training-validation success (82.03\%), and constructs graphs from eight rollouts for each of 140 ALFWorld tasks. Candidate actions must be legal and share the exact simulator-state signature, student-prompt hash, and remaining budget. The sampling protocol requires an absolute graph-score difference above $10^{-8}$ and positive recorded reliability for at least one source visit. The latter requires that its state--action pair occur in at least two trajectories, have nonzero empirical advantage, and appear in at least 16 of 64 trajectory-bootstrap replicates. Each replicate resamples the eight trajectories with replacement. These criteria select the evaluation pool; allocation uses the objective and pair weights defined in Appendix~\ref{app:implementation}.

The pool contains 311 action pairs across 94 tasks. Using selection seed 20261027, we choose one pair uniformly within each task, reserve eight disjoint tasks for replay checks, and select 60 formal tasks proportionally across the remaining task categories. The state and action manifest is frozen before continuation sampling. For each pair, we restore the recorded state and history, force either action, and sample 32 continuations per action with the remaining interaction budget. Sampling uses temperature 1, top-$p$ 1, no top-$k$ restriction, and at most 512 response tokens per turn. Request seeds differ between graph construction and continuation evaluation; the inference-engine initialization seed is 0. The resulting 3,840 continuations are excluded from graph estimation and pair selection.

\textbf{Outcome metrics.}
For each action, we average the independent discounted returns $\sum_{j\geq0}\gamma^j r_j$, with $\gamma=0.1$, binary terminal success reward, and $j=0$ denoting the forced action. Ranking agreement compares a predicted gap's sign with the difference between these two sample means. All 60 empirical return differences exceed the outcome sign tolerance $10^{-12}$. Graph gaps use tolerance $10^{-8}$; teacher and credit gaps use $10^{-6}$. Success is also measured without discounting. Table~\ref{tab:independent-progress-detail} reports both endpoints. Proportion intervals are 95\% Wilson intervals; paired differences use 10,000 task-level bootstrap resamples with seed 20261027, preserving the two actions within each task.

\begin{table}[t]
\centering
\footnotesize
\setlength{\tabcolsep}{3pt}
\begin{tabular*}{\columnwidth}{@{\extracolsep{\fill}}lrr@{}}
\toprule
Measure & Estimate & 95\% interval \\
\midrule
Teacher return agreement & 50.0\% & [37.7, 62.3] \\
Graph return agreement & 81.7\% & [70.1, 89.4] \\
Mean return difference & $+0.03461$ & [0.00054, 0.08394] \\
Success difference (pp) & $+2.81$ & [0.05, 5.89] \\
\bottomrule
\end{tabular*}
\caption{Independent outcomes for 60 matched action pairs. Differences subtract the alternative action from the graph-preferred action.}
\label{tab:independent-progress-detail}
\end{table}

\textbf{Selective calibration during training.}
The Subgoal Progress Graph supplies direct action comparisons across 38.35\% of sampled trajectories while concentrating them on 1.99\% of turns (Table~\ref{tab:training-pair-coverage}). We measure final eligible comparisons during SGR-OPSD training on ALFWorld with the 3B model and seed0, using empirical Q--V values and pairwise weight calibration. Across 150 updates, 6,539 pairs enter completed policy updates; trajectory coverage rises from 30.97\% in updates 1--50 to 45.80\% in updates 101--150. 

\begin{table}[t]
\centering
\footnotesize
\setlength{\tabcolsep}{3pt}
\begin{tabular*}{\columnwidth}{@{\extracolsep{\fill}}lrr@{}}
\toprule
Metric & Count / total & Fraction \\
\midrule
Direct task coverage & 947 / 2,396 & 39.52\% \\
Direct trajectory coverage & 7,364 / 19,200 & 38.35\% \\
Direct turn coverage & 13,878 / 698,343 & 1.99\% \\
\midrule
Minimum matched support = 1 & 5,545 / 6,539 & 84.80\% \\
Minimum value-pool support = 1 & 5,545 / 6,539 & 84.80\% \\
\bottomrule
\end{tabular*}
\caption{Pairwise calibration coverage and support counts during ALFWorld 3B training (seed0). Coverage aggregates per-update counts; support counts distinct trajectories per action and takes the minimum across the pair.}
\label{tab:training-pair-coverage}
\end{table}

\textbf{Preference reliability at low support.}
Matched support counts visits satisfying the comparison conditions; value-pool support counts trajectories contributing to the original graph node--action estimate. In the separate 60-task independent verification set, 46 pairs have minimum value-pool support of one. Subgoal Progress Graph preferences agree with independent returns on 40/46 (87.0\%), versus 21/46 (45.7\%) for teacher evidence. Among the 15 directly eligible pairs in this group, calibration improves credit agreement from 6/15 (40.0\%) to 12/15 (80.0\%). For the remaining 14 pairs with support of at least two, graph and teacher agreement are both 9/14; calibration improves credit agreement from 2/6 to 4/6 on the eligible subset. These comparisons assess preference and credit reliability by support level using the same independent continuations.

\textbf{Coverage and state merging.}
Candidate comparisons cover 67.1\% of tasks and 58.8\% of trajectories in the graph-building pool, with direct participation concentrated in 3.95\% of turns (Table~\ref{tab:progress-coverage}). Coverage counts each task, trajectory, or turn once before nonzero-advantage allocation filtering. This measures where action comparisons are available; the trajectory budget determines how their calibration also affects other turns.

\begin{table}[t]
\centering
\footnotesize
\setlength{\tabcolsep}{3pt}
\begin{tabular*}{\columnwidth}{@{\extracolsep{\fill}}lrr@{}}
\toprule
Covered unit & Count / total & Fraction \\
\midrule
Tasks & 94 / 140 & 67.1\% \\
Trajectories & 658 / 1,120 & 58.8\% \\
Turns & 1,055 / 26,717 & 3.95\% \\
\bottomrule
\end{tabular*}
\vspace{5pt}
\begin{tabular*}{\columnwidth}{@{\extracolsep{\fill}}lrrr@{}}
\toprule
Additional node key & Same & Reversed & Below threshold \\
\midrule
\multicolumn{4}{@{}l}{\textit{Formal pairs ($n=60$)}} \\
Remaining budget & 52 & 2 & 6 \\
Budget and prompt & 52 & 2 & 6 \\
\midrule
\multicolumn{4}{@{}l}{\textit{All candidates ($n=311$)}} \\
Remaining budget & 278 & 8 & 25 \\
Budget and prompt & 278 & 8 & 25 \\
\bottomrule
\end{tabular*}
\caption{Candidate coverage and sensitivity to node splitting. Directions are relative to the original graph; the gap threshold is $10^{-8}$.}
\label{tab:progress-coverage}
\end{table}

Splitting nodes by remaining budget preserves 52 of the 60 evaluated preference directions, reverses two, and leaves six below threshold. Adding prompt identity produces no further direction changes. Both refinements agree with independent returns on 44/54 remaining predictions (81.5\%), with six abstentions. The diagnostic holds candidate actions and independent outcomes fixed while recomputing graph values under each node key. The original nodes include physical-state signatures; among 17,991 visited source nodes, 4,375 have repeated visits, 3,513 mix remaining budgets, and 3,540 mix student-prompt hashes. These mixed nodes account for 9,440 and 9,502 of the 26,717 turns, respectively. The results quantify how successor pooling affects preferences even when the compared actions share a matched starting state.

For example, the original graph favors \texttt{go to countertop 1} over \texttt{go to cabinet 1} by $0.000333$; budget splitting reverses the gap to $-0.000045$. Both actions succeed in all 32 continuations, while the countertop action has $0.00028884$ higher mean discounted return. This pair illustrates why success and discounted return are reported separately, and why finer merging keys are assessed against independent outcomes.

\begin{table}[t]
\centering
\footnotesize
\setlength{\tabcolsep}{3pt}
\begin{tabular*}{\columnwidth}{@{\extracolsep{\fill}}lrrr@{}}
\toprule
Signal & Correct & Incorrect & Tied \\
\midrule
\multicolumn{4}{@{}l}{\textit{All 60 pairs}} \\
Teacher evidence & 30 & 30 & 0 \\
Graph preference & 49 & 11 & 0 \\
Teacher-reference credit & 11 & 14 & 35 \\
Calibrated credit & 19 & 6 & 35 \\
\midrule
\multicolumn{4}{@{}l}{\textit{Same 21 directly eligible pairs}} \\
Teacher evidence & 9 & 12 & 0 \\
Graph preference & 16 & 5 & 0 \\
Teacher-reference credit & 8 & 13 & 0 \\
Calibrated credit & 16 & 5 & 0 \\
\bottomrule
\end{tabular*}
\caption{Credit-ranking agreement with independent mean continuation returns. Correct, incorrect, and tied predictions form disjoint categories.}
\label{tab:credit-ranking-complete}
\end{table}

\subsection{Credit Calibration Analysis}
\label{app:credit-calibration-analysis}
\textbf{Improved outcome alignment.}
Calibration using Subgoal Progress Graph preferences increases agreement with independent returns from 8/21 (38.1\%) to 16/21 (76.2\%) on the same directly eligible pairs, with Wilson intervals [20.8\%, 59.1\%] and [54.9\%, 89.4\%] (Table~\ref{tab:credit-ranking-complete}). Across all 60 pairs, calibration yields eight additional correct rankings: nine corrections and one change from correct to incorrect. The paired gain is 13.3 percentage points over all 60 pairs (95\% task-bootstrap interval [3.3, 23.3]). Thirty-five pairs involve zero-advantage visits and remain tied under both reference and calibrated credit; four further pairs fall outside the exact direct-comparison conditions. The table reports correctness and ties for both the complete set and the directly eligible subset.

\textbf{Online scoring and allocation.}
We analyze teacher-reference and calibrated credit on recorded rollout traces for the 60 formal tasks. We save  teacher-reference and calibrated weights  during training. This controlled comparison evaluates the allocation mechanism on the selected rollouts. Teacher and student score the same saved response under their respective contexts. 

The traces includes all eight graph-building trajectories per task: 8,202 unique turns, of which 8,201 have scoreable actions. Both actions in every selected pair are scoreable. Base advantages use the frozen GRPO reward normalization and invalid-action penalty; token masses count the canonical response tokens once per unique visit. We use $\epsilon=0.1$, $T=T_G=1$, $\beta=2$, $\mu=0.02$, and graph-score bound $b_G=0.3$, with the solver checks in Appendix~\ref{app:implementation}.

For pair $e$, raw teacher evidence is $b_e^\top d$, while teacher-reference and calibrated credit gaps are $b_e^\top c^{\mathrm{ref}}$ and $b_e^\top c^*$, respectively, with $c^{\mathrm{ref}}=s\odot(w^{\mathrm{ref}}-\mathbf1)$. These sign-aware gaps compare favorable credit across advantage signs. We select 21 eligible pairs based on action scoreability, nonzero advantages, and matched states, contexts, and remaining budgets across their source visits, independently of continuation outcomes.

\subsection{Sensitivity to the Evaluation Discount}
\label{app:discount-sensitivity}
Preferences from the Subgoal Progress Graph outperform teacher evidence at every tested evaluation discount, and calibrated credit improves on the teacher reference throughout (Table~\ref{tab:discount-sensitivity}). We reuse all 3,840 independent continuations, varying their return discount $\gamma_{\mathrm{eval}}$ while keeping the graph, selected pairs, and all predicted rankings fixed. Graph agreement is 81.7\% for $\gamma_{\mathrm{eval}}=0.1$--$0.7$ and 76.7\% at $0.9$; credit agreement rises from 38.1\% to 76.2\% and from 47.6\% to 66.7\%, respectively. At $\gamma_{\mathrm{eval}}=1$, returns equal binary success, and both comparisons retain their advantage on non-tied pairs. This analysis evaluates fixed predictions under different return objectives.

\begin{table}[t]
\centering
\footnotesize
\setlength{\tabcolsep}{3pt}
\begin{tabular*}{\columnwidth}{@{\extracolsep{\fill}}ccccc@{}}
\toprule
& \multicolumn{2}{c}{Action preference} & \multicolumn{2}{c}{Credit allocation} \\
\cmidrule(lr){2-3}\cmidrule(lr){4-5}
$\gamma_{\mathrm{eval}}$ & Teacher & Graph & Reference & Calibrated \\
\midrule
0.1 & 30/60 & \textbf{49/60} & 8/21 & \textbf{16/21} \\
0.3 & 28/60 & \textbf{49/60} & 8/21 & \textbf{16/21} \\
0.5 & 28/60 & \textbf{49/60} & 8/21 & \textbf{16/21} \\
0.7 & 28/60 & \textbf{49/60} & 8/21 & \textbf{16/21} \\
0.9 & 29/60 & \textbf{46/60} & 10/21 & \textbf{14/21} \\
1.0 & 16/31 & \textbf{22/31} & 10/17 & \textbf{12/17} \\
\bottomrule
\end{tabular*}
\caption{Agreement with independent returns across evaluation discounts on ALFWorld (3B). Entries are correct/non-tied pairs. At $\gamma_{\mathrm{eval}}=1$, 29/60 action pairs and 4/21 credit pairs have tied reference returns and are excluded from the denominators.}
\label{tab:discount-sensitivity}
\end{table}

\end{document}